\documentclass[twoside,11pt]{article}
\usepackage{fullpage}

\usepackage{amsmath,amsfonts,bm}

\def\eqref#1{equation~\ref{#1}}

\def\1{\bm{1}}

\DeclareMathAlphabet{\mathsfit}{\encodingdefault}{\sfdefault}{m}{sl}
\SetMathAlphabet{\mathsfit}{bold}{\encodingdefault}{\sfdefault}{bx}{n}

\usepackage{graphicx} 
\usepackage{microtype}
\usepackage{subcaption}
\usepackage{booktabs}
\usepackage{hyperref}
\usepackage{natbib}
\usepackage{lastpage}

\date{}

\begin{document}

\title{Is Pre-Training Truly Better than Meta-Learning?}
\author{Brando Miranda\thanks{Department of Computer Science, Stanford University. Equal Contribution}, Saumya Goyal\thanks{Machine Learning Department, Carnegie Mellon University. Equal Contribution}, Patrick Yu\thanks{Department of Computer Science, University of Illinois Urbana-Champaign}, Yu-Xiong Wang\footnotemark[3], Sanmi Koyejo\thanks{Department of Computer Science, Stanford University}}

\maketitle

\begin{abstract}
In the context of few-shot learning, it is currently believed that pre-training using a fixed pre-trained model, along with fine-tuning the final layer during evaluation, outperforms standard meta-learning algorithms. 
We re-evaluate these claims under an in-depth empirical examination of an extensive set of diverse datasets and compare baseline pre-training and meta-learning algorithms. 
Unlike previous work, we emphasize a fair comparison by; using the same architecture, using only model-agnostic algorithms for both techniques, and training to convergence. 
We then use a previously proposed metric -- the diversity coefficient -- to compute the average quantitative diversity of a dataset.
Using this analysis, we demonstrate the following:
(i) when the diversity of a dataset is low, pre-training beats meta-learning on average, and 
(ii) when the diversity is high, meta-learning beats pre-training on average
(iii) however, overall, there is no statistical difference in the two methods
Our extensive experiments consider 21 few-shot learning benchmarks on vision, including the large-scale few-shot learning dataset Meta-Dataset and train 196 models to convergence. 
These
observations are contrary to the currently held belief that pre-training is always better than meta-learning.
\end{abstract}

\section{Introduction}
Research in Artificial Intelligence (AI) has been significantly influenced by the goal of learning from a few examples. One leading approach is meta-learning, embodying the notion of "\textit{learning to learn}" or "\textit{learning to adapt}" \citep{meta_learning}.
Notably, these approaches have been successfully combined with deep learning in computer vision \citep{alexnet, resnet}, natural language processing \citep{bert, gpt3}, game playing \citep{alphago, atari1, EfficientZero}, theorem proving \citep{skiptree, gptf, pact}, code \citep{codex}.
Few-shot learning challenges a model's ability to learn a new task from a few examples quickly and, therefore, has been a prominent research area for applying meta-learning algorithms.
However, recent findings \citep{rfs} have demonstrated that a pre-trained model with a fixed embedding can surpass many sophisticated meta-learning algorithms in performance on a variety of few-shot learning benchmarks \citep{rfs, Chen2019, Chen, Dhillon2019, Huang2019}.
It has been suggested that we need to ``Re-think Few-Shot Image Classification; is a Good Embedding All You Need?" Thus, the emerging consensus is that transfer learning with pre-training uniformly outperformed meta-learning. We challenge the views presented in these works and re-evaluate model-agnostic methods for meta-learning and transfer learning via pre-training. Our main result is that a more nuanced analysis shows that data diversity plays a key role, which was previously unknown, in interpreting the comparison across methods.
\\
Our \textbf{key contributions} are as follows:
\begin{enumerate}
    \item We present a novel re-evaluation that employs a fair comparison by using only model-agnostic algorithms for both techniques and training to convergence. Additionally, it only uses fundamental and comparable algorithms for both techniques to avoid confirmation biases from more extensive work in a particular area.
    Specifically, we compare pre-training with head-only fine-tuning (PT) against the Model Agnostic Meta-Learning (MAML) algorithm.
    \item We evaluate differences using statistical hypothesis testing -- in particular, we use the effect size as measured by Cohen's d~\citep{Cohen_1992} to characterize more carefully the difference in performance between the two methods.
    \item We do our analysis from a novel data-centric perspective by measuring the diversity of the datasets.
    This reveals an additional rich structure to our findings. We find that:
    \begin{enumerate}
        \item When the diversity of a dataset is low, pre-training beats meta-learning on average 
        \item When the diversity of a dataset is high, meta learning beats pre-training 
        \item When all dataset diversities are considered, the effect size is close to zero
        , favoring neither algorithm.
    \end{enumerate}
\end{enumerate}

\section{Background and Preliminaries}\label{background}
\textbf{Few-shot learning:}
Few-shot learning involves training a model to perform a task using very few examples of the task. Since this is infeasible for large models and complicated tasks, we generally seek to leverage information from examples of other, related tasks. 

In this paper, each experiment uses a large corpus of classification data consisting of inputs with $N$ different labels. Each task is a classification task between a $n(<N)$ labels from the corpus (unless stated otherwise). \\
\textbf{Transfer Learning via Pre-Training (PT):}
A common technique for few-shot learning of a task $\tau_A$ is to train a model on a task (or set of tasks) $\tau_B$ that is similar to $\tau_A$ but for which we have a lot more examples. We can then fine-tune the model using the few examples we have of $A$.
Prior research \citep{rfs} demonstrates that an initialization pre-trained with a union of all data from all labels can supersede numerous meta-learning methods. 
Specifically, their methodology involves two phases: initially, they utilize a union of all labels and undertake meta-training with conventional pre-training (PT).
Subsequently, during the meta-testing phase, they employ a standard inference method prevalent in transfer learning via pre-training: extraction of a fixed feature from the neural network and a comprehensive fine-tuning of the final classification layer (the head) with LBFGS (Limited-memory Broyden–Fletcher–Goldfarb–Shannon algorithm).\\
\textbf{Model-Agnostic Meta-Learning (MAML):}
The MAML algorithm \citep{maml} is designed to meta-learn an optimal initialization of neural network parameters, thereby priming it for rapid gradient descent adaptation for few-shot learning.
The algorithm works using a large set of few-shot learning tasks for training (or meta-training) and consists of two core optimization loops. 
The inner loop executes fast adaptation based on the few-shot learning task used for meta-training in the corresponding outer loop iteration. 
The outer loop iterates through tasks and primes the parameters for swift adaptation, so that the resulting network after inner-loop optimisation performs well on the task.
During meta-testing/evaluation, the inner loop exclusively carries out the adaptation of the representation acquired from the outer loop. 
For this paper, all the few-shot tasks for each experiment  are $n$-way $k$-shot learning tasks sampled from the labelled corpus corresponding to that experiment (unless stated otherwise).\\
\textbf{Effect Size (Cohen's d):}
Cohen's d \citep{Cohen_1992} is a statistical tool designed to quantify the size or magnitude of an effect, irrespective of the sample size. 
This standardized measure of effect size allows for the comparison of results across different studies and domains. 
It is calculated by determining the difference between two means and dividing by the pooled standard deviation (approximately unbiased estimate of the combined standard deviations),
providing a measure of the effect size in terms of standard deviation units: 
$d = (\mu_1 - \mu_2) / pooled\_std(\sigma_1, \sigma_2)$.
We explain the main reason for using this metric and its corresponding decision rule in Section \ref{methods}.\\
\textbf{Task2Vec Embeddings of tasks:}
We use the Task2Vec diversity coefficient proposed in \cite{curse_low_div} to compute the diversity of a dataset (or a few-shot learning benchmark).
To understand the diversity coefficient, we explain how to compute Task2Vec (vectorial) embeddings of a task and briefly explain why it is a good vectorial embedding of a task.  
Task2Vec \citep{task2vec} embeds data (e.g., any batch) using the diagonal entries of the Fisher Information Matrix (FIM) using a fixed neural network (also called a \textbf{probe network}) after (partially) fine-tuning the final layer to solve the current task (or batch).
Thus, the Task2Vec embedding of task $\tau$ is:
\begin{equation}\label{fim}
    \vec{f}_{D_\tau, f_w} = Diag(\hat F_{D_\tau, f_w})
\end{equation}
where $Diag$ extracts the diagonal of the FIM:
\begin{equation*}
    \hat F_{D_\tau, f_w} = \mathbb E_{x, y} \nabla_w \log p(y \mid x, f_w) \nabla_w p(y \mid x, f_w)^{\top}
\end{equation*}
and $f_w$ is the fixed probe network with architecture $f$ with weights $w$, 
$x$ is sampled from the batch/data $D_{\tau} = \{ (x_i, y_i ) \}^n_{i=1}$ for task $\tau$,
$y$ is sampled from the (empirical) posterior distribution using the probe network i.e., $p(y\mid x, f_w)$.
This is a good embedding of tasks because the (diagonal) of the FIM indicates the most informative weights for solving the current task and thus serves as a unique fingerprint for task distribution. 
The Task2Vec authors \citep{task2vec} empirically validate their embeddings, e.g., Task2Vec embeddings cluster in a way that matches human semantic relations between different visual tasks \citep{task2vec} and Task2Vec yields (cosine) distance that positively correlates with taxonomical distances \citep{task2vec}.\\
\textbf{Task2Vec Diversity coefficient:}
The Task2Vec diversity coefficient is a quantitative metric proposed by \cite{curse_low_div} to approximate the effective number of tasks in a dataset.
If the tasks are probability distributions, then this metric approximates the average distance between probability distributions.
They validate it with synthetic experiments where the ground truth diversity is known.
We further validate it in the supplementary section \ref{div_coeff}.
We show the intuitive notion that when different types of datasets are unioned to create a new dataset, the Task2Vec diversity coefficient increases.
The \textit{Task2Vec diversity coefficient} is defined as the expected (cosine) distance between Task2Vec embeddings of different tasks (or data batches) for a fixed probe network from a few-shot learning benchmark/dataset $B$:
\begin{equation}\label{true_diversity_coeff1}
    \hat div (B) = 
        {\mathbb E}_{\tau_1, \tau_2}
            {\mathbb E}_{D_1, D_2} 
                d(\vec{f}_{D_1, f_w}, \vec{f}_{D_2, f_w} ) 
\end{equation}
where $F_{D_1, f_w}$ is the Task2Vec embedding of the training data $D_{\tau} = \{ (x_i, y_i ) \}^n_{i=1}$ for task $\hat p(x, y \mid \tau)$ that uses the fixed probe network $f_w$ with architecture $f$ and weights $w$.
Note that $\tau_1, \tau_2$ are tasks sampled from the (meta) distribution of tasks $\hat p(\tau \mid B)$ for the current benchmark $B$, and $d$ is the cosine distance.
It is worth restating that in this framework, a task is defined as an n-way, k-shot few-shot learning task.
Each task thereby includes $n$ classes, each of which is sampled with $k$ examples.
Usually, these $k$ examples form the support (train) set for fast adaptation and we also sample a query (test) set for evaluation.

\section{Statistical Methods}\label{methods}
In this section, we explain the main statistical methodology we use to analyze our results which is the effect size (Cohen's d) \citep{Cohen_1992, andrade2020meandifference} described in Section \ref{background} and its rationale.

\subsection{Decision Rule}
In addition to using raw effect size values to determine impact of the two training techniques, we also devise a decision rule in accordance with typical machine learning literature.
\\
\textbf{Summary of decision rule: } 
To determine if there was a significant difference between pre-training and meta learning we computed the meta-test accuracy of the two methods and determined if the effect size was larger than what is common in machine learning (we use a 1 \% standardized threshold by dividing by the pooled standard deviation \citep{rodríguez2020embedding, li2021universal}). 
If it is larger, we determine there is a significant difference and report which method is better by looking at the sign of the effect size.
Otherwise, if the effect size was small according to this threshold, we reported there was no significant difference.
All our tables include the raw effect size values. 

More precisely, first compute a list of (meta) test accuracies on a batch of few-shot learning tasks for meta learning and pre-training denoted as: $accs_{meta}$ and $accs_{pt}$.
Compute the effect size (ES) of these and compute the threshold $\delta_{1\%}$ by dividing 1 \% by the pooled standard deviation of the current test accuracies, i.e. divide $\delta_{1\%} = \frac{0.01}{pooled\_std(accs_{pt}, accs_{meta})}$.
Then use the decision rule in equation \ref{decision_rule} to determine whether to reject the null hypothesis or not.
\begin{figure*}
\begin{equation}\label{decision_rule}
  Decision(accs_{pt}, accs_{meta}) =
    \begin{cases}
      \text{H0 (no diff.)} & \text{if } ES(accs_{pt}, accs_{meta}) \in [-\delta_{1\%}, \delta_{1\%}] \\
      \text{H1 (pt)} & \text{if } ES(accs_{pt}, accs_{meta}) > \delta_{1\%} \\
      \text{H1 (meta)} & \text{if } ES(accs_{pt}, accs_{meta}) < -\delta_{1\%}
    \end{cases}
\end{equation}
\end{figure*}

Note, you can also look at the effect size sign to determine which method performed best.
If the sign is positive, then pre-training outperformed meta learning, otherwise meta learning performed better.
i.e. there wasn't a significant difference between the methods.
We refer the reader to the Appendix \ref{sec:es_just} for more discussion of the decision rule.

\section{Experiments}\label{experiments}
We compare the performance of a transfer learned model against a meta learned model using the effect size and the decision procedure described in Section \ref{methods}.
To provide further insights, we organize our experiments results according to the diversity of the datasets.
In particular, we divide the results into datasets with low and high diversity.
We set the division of low vs high at $0.146$ Task2Vec diversity, because that was approximately the average diversity.
In addition, this division roughly divided the datasets in an intuitive way: most datasets that were a union of four or more datasets had a diversity higher than $0.146$, while the rest were below.
The only exception was Omniglot, which had a high diversity, but also contains a vast amount of visual concepts (over 1000).
We present our experiment details in Section \ref{exp_detail}, an overview of our results in Section \ref{experiments_summary}, and refer the reader to Sections \ref{sec:exp_details} and \ref{datasets} for detailed results and explanations of datasets.

\subsection{Experimental Details}\label{exp_detail}
In order to provide a comprehensive assessment of our claims, we train several different architectures including Resnet12, Resnet50 and hand-crafted CNNs with several different filter sizes. We train these models on a total of 21 few-shot learning benchmarks, including the large-scale few-shot learning dataset Meta-Dataset \citep{Triantafillou2019}. Each model is trained using PT for transfer learning via pre-training, as well as MAML for meta-learning with 5 and 10 inner loops (MAML5 and MAML10 respectively) as described in Section \ref{background} yielding a total of 196 models trained to convergence. For training, all few-shot learning tasks are devised as 5-shot 5-way tasks with 15 examples for evaluation. 

We refer the reader to the Appendix \ref{sec:exp_details} for detailed hyperparameter details and results for each model.

\subsection{Overview of Results} \label{experiments_summary}
Tables \ref{tab:low_div_summary} and \ref{tab:high_div_summary} show the effect sizes of Resnet models trained with Pre-training (PT) vs. first-order (FO) MAML on popular low and high diversity datasets respectively.
In the low diversity setting pre-training outperforms meta learning but, in the high diversity setting, meta learning outperforms pre-training.
In the low diversity setting, pre-training has a larger effect size relative to the effect size of meta learning in the high diversity regime when analyzing the results in this way. 

\begin{figure}
    \centering
    \begin{subfigure}{0.5\textwidth}
        \centering
        \includegraphics[width=\linewidth]{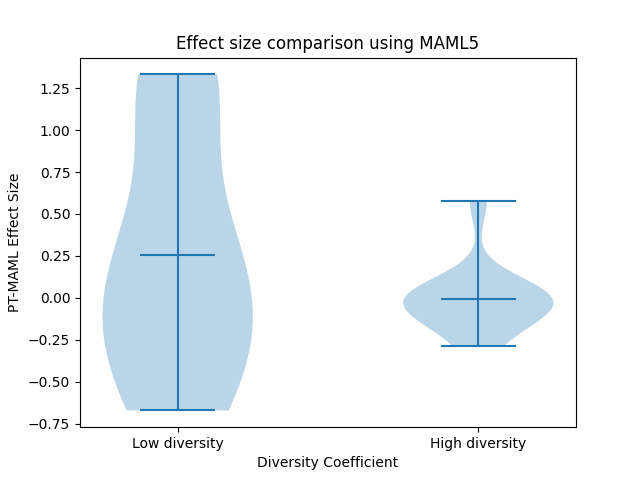}
        \caption{Comparison of effect sizes running MAML5}
    \end{subfigure}%
    \begin{subfigure}{0.5\textwidth}
        \centering
        \includegraphics[width=\linewidth]{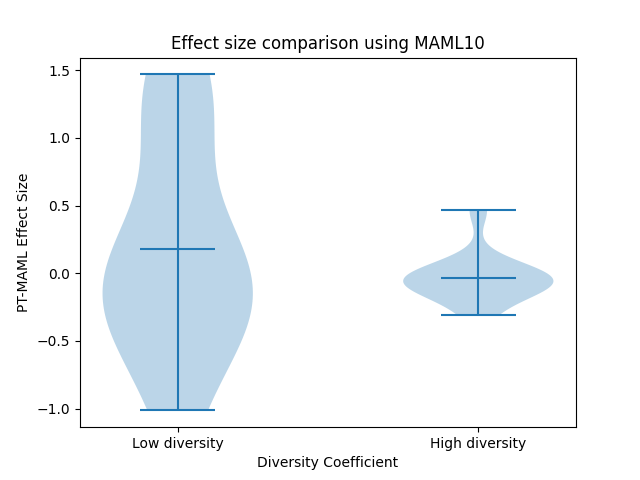}
        \caption{Comparison of effect sizes running MAML10}
    \end{subfigure}
    \caption{\textbf{Trend of observed effect sizes with diversity.} Here MAML5 and MAML10 represent running 5 inner adaptation steps and 10 inner adaptation steps respectively.}
    \label{fig:es_trends}
\end{figure}

In Figure \ref{fig:es_trends} we show the overall trend of effect sizes observed for low and high diversity datasets using MAML5 and MAML10.
In Table \ref{tab:summary_avg_es_low_vs_high_divs} we summarize all results and compare the low diversity vs the high diversity regime when there is a significant difference (i.e. H1).
In more detail, when averaging all the low diversity results with an H1 (pt or meta) decision, the overall effect size is 0.103, favoring pre-training.
When averaging all the high diversity results, the overall effect size is -0.107 favoring meta learning.
However, note that both of these are considered low effect in classical statistics, as the regime for small, medium, high are roughly 0.2, 0.5, 0.8 \citep{andrade2020meandifference, Cohen_1992}.

\begin{table*}[h!]
\centering
\caption{
\textbf{Effect size between a Pre-trained (PT) model vs. first-order (FO) MAML model on low diversity few-shot learning vision datasets.}
Here the null hypothesis (H0) refer to the cases when both meta learning and pre-training perform equivalently according to our decision rule. H1 (pt) and H1 (meta) refer to the cases where pre-training and meta learning performed better respectively.
}
\begin{tabular}{|c|c|c|}
\hline
Dataset (model) & ES (PT vs MAML5) & ES (PT vs MAML10) \\ \hline

CIFAR-FS (Resnet12) & -0.266 (H1 meta) & -0.342 (H1 meta) \\
FC100 (Resnet12) & -0.251 (H1 meta) & -0.248 (H1 meta) \\
mini-ImageNet (Resnet12) & 0.413 (H1 pt) & 0.149 (H1 pt) \\
tiered-ImageNet (Resnet12) & 0.218 (H1 pt) & 0.0290 (H0 no diff.) \\
\hline
\end{tabular}
\label{tab:low_div_summary}
\end{table*}

\begin{table*}[h!]
\centering
\caption{
\textbf{
Effect size between a Pre-trained (PT) model vs MAML model on high diversity few-shot learning vision datasets.} 
Here the null hypothesis (H0) refer to the cases when both meta learning and pre-training perform equivalently according to our decision rule. H1 (pt) and H1 (meta) refer to the cases where pre-training and meta learning performed better respectively.
}
\begin{tabular}{|c|c|c|}
\hline
Dataset (model) & ES (PT vs MAML5) & ES (PT vs MAML10) \\ \hline

Omniglot (Resnet12) & 0.00702 (H0 no diff.) & 0.0679 (H0 no diff.) \\

MIO (Resnet12) & -0.0197 (H0 no diff.) & -0.0161 (H0 no diff.) \\

MDS (Resnet50, avg over 4 seeds) & 0.001375 (H0 no diff.) & -0.064275 (H1 meta) \\
\hline
\end{tabular}
\label{tab:high_div_summary}
\end{table*}


\begin{table*}[h]
\centering
\caption{\textbf{Average effect size (Cohen's d) for each statistical decision rule comparing pre-training and meta learning.}
Here the null hypothesis (H0) refer to the cases when both meta learning and pre-training perform equivalently according to our decision rule. H1 (pt) and H1 (meta) refer to the cases where pre-training and meta learning performed better respectively.
}
\begin{tabular}{|c|c|c|c|}
\hline
Setting & Dec: H0 & Dec: H1 (pt) & Dec: H1 (meta) \\ 
\hline
Low diversity & 0.029 & 0.7292 & -0.556 \\
High diversity & 0.0521 & 0.0717 & -0.164 \\
\hline
\end{tabular}
\label{tab:summary_avg_es_low_vs_high_divs}
\end{table*}



\section{Related Work}
Our approach aims to develop a data centric-oriented framework for examining meta-learning algorithms in response to the call to re-think meta-learning -- especially in the context of few-shot learning \citep{rfs}.
Also, the main distinction between our work and theirs \citep{rfs} lies in:
that we bring clarity and nuance to their surprising results using our focus on the diversity of datasets, and 
that we use a much wider set of diverse datasets.
In addition, unlike their work, ours focuses in ensuring a fair comparison by using a consistent neural network architecture, optimizer, and all models trained to convergence. 
Another point of difference is their attainment of further accuracy gains through distillation, a method we have not analyzed but will consider for future work.

Some earlier work demonstrated that MAML operates chiefly through feature reuse \citep{Raghu2020} rather than rapid learning, signifying that a model trained with MAML undergoes minimal alteration after the MAML adaptation. 
Our work diverges from theirs primarily in two ways: 
1) we contrast MAML meta-trained models against models that are pre-trained with a union of all the data rather than solely comparing varying types of MAML models, and 
2) we contextualize our analysis by explicitly analyzing  properties of datasets like diversity. 

Related work also includes the predictability of adversarial transferability and pre-training through extensive experimentation and a theoretical analysis \citep{Liang2021}. 
The primary difference between their work and ours is their primary focus on pre-training, while we concentrate on meta-learning for few-shot learning. 
Additionally, we did not consider adversarial transferability, which forms a central part of their analysis.

Now we comment on datasets/benchmarks related to our work.
The meta-dataset benchmark aims to create a larger and more diverse dataset for few-shot learning \citep{Triantafillou2019}.
The key distinction between their work and ours is our use of a quantitative metric to measure the intrinsic diversity of a dataset, ad therefore bring more nuanced considerations beyond dataset size or even just class numbers. 
Another interesting work is the IBM Cross-Domain few-shot learning benchmark \citep{bscd_fsl}.
They provide an interesting benchmark, but 
cross-domain learning is out of scope for our work.

The work by \cite{global_labels} proposes the concept of global labels, equivalent to what we call pre-training in this paper. 
Their theoretical analysis, however, is dependent on a fixed feature extractor, and fails to accommodate different feature extractors e.g. when comparing pre-training vs MAML.

\section{Discussion}
Our paper presents an alternative explanation for claims that a model trained with pre-training can often beat a model trained with meta-learning \citep{rfs, Chen2019, Chen, Dhillon2019, Huang2019}. 
In particular, we show that a more careful analysis -- especially one that takes the dataset diversity into account -- can provide a nuanced conclusion.
Note, however, all statistical tools have assumptions and none are perfect -- as the following discussion will exemplify.

For example, Section \ref{experiments_summary} shows that when the overall average effect size across all high diversity datasets is used, we get a mean effect size of -0.107, favoring meta learning.
Similarly, across all low diversity datasets the effect size is 0.103, favoring pre-training.
However, this effect size is considered low in classical statistics, since effect sizes of small, medium, high are roughly 0.2, 0.5, 0.8 \citep{Cohen_1992}.
A final nuance we'd like to add to our results is the variance of the effect size. 
We note the effect sizes ranges from 0.778 to -0.717. 
Meaning that when looking at individual experiments, the difference might be high for each individual run.

In addition, we want to emphasize that unlike previous work, we truly emphasized a fair comparison i.e. we used the same architecture, comparable and model-agnostic algorithms and trained all models to convergence.
Previous results, especially \cite{rfs}, compared different architectures and different meta-learning methods, thereby making it impossible to know the true source of improved performance. 
We disambiguate this and show that in datasets with low diversity, pre-training outperforms meta learning, while meta learning outperforms pre-training in datasets with high diversity. In addition, when all dataset diversities are considered, neither method is better. We additionally demonstrate this result for language models as detailed in Appendix \ref{sec:gpt}.

We'd like to note that our low diversity results are in conflict with previous results in \citep{curse_low_div}.
We conjecture the difference was mainly due to:
(i) The batch size they used in conjunction with their use of Confidence Intervals (CIs) as their statistical analysis. 
Their experiments had (meta) batch size of 100, leading to overly large intersecting CIs that leads them to conclude pre-training and MAML are equivalent.
(ii) The number of models they trained is less than half of ours. 
We trained 40 models in total for the low diversity setting, and they did 18 (less than half of ours).

We'd also like to note that although MAML and PT are marginally different in performance, MAML is still harder to train than PT. 
In particular, MAML requires an additional memory for the gradient in the forward pass and makes it harder to train for large models.
Even though \citep{bronskill2021memory} memory efficient meta-learning might solve the memory issue, it is still less simple than pre-training.

In contrast to \cite{rfs}, our work deliberately opts for a broad suite of datasets with varying diversities at the expense of exploring fewer meta-learning methods, which might be viewed as a potential limitation. 
This choice arises from \cite{rfs}'s implication that pre-training may outperform virtually all meta-learning algorithms (they considered).
Our research, however, challenges this view \citep{rfs}, and demonstrates that even the simplest meta-learning algorithm, MAML, can outperform pre-training when the comparison employs fair and rigorous statistical analysis that considers the characteristics of the dataset like the diversity. 
In other words, we only need a single example (i.e., a single meta-learning algorithm) to provide a proof by counter example.
Our approach underscores the importance of dataset properties on algorithm performance, thereby contributing a nuanced perspective to the ongoing discourse in the meta-learning domain.

The work by \cite{Kumar2022} provides an exploration of the effects of diversity in meta-learning. 
However, they focus mostly on sampling strategies, while we focused on the intrinsic diversity in the datasets/benchmarks themselves.

\bibliography{refs}
\bibliographystyle{plainnat}

\newpage

\appendix
\section{Supplementary Material}

\subsection{Related Work (Cont.)}


The study by \cite{Chen2021} offers, to our knowledge, the first non-vacuous generalization bounds for the (supervised) meta-learning setting. 
However, their results do not aim to differentiate classes of meta-learning, as our work attempts to do empirically. 


The study by \cite{Denevi2020} presents a theoretical treatment of meta-learning using meta-learners with closed-form equations derived from ridged regularization using fixed features. However, their results are theoretical and they do not explore their findings to modern few-shot learning benchmarks like MiniImagenet, Cifar-fs, FC100, TieredImagenet, Meta-Dataset, etc. like we do.



The work by \cite{Rosenfeld2021} provides a theoretical analysis on the difference between interpolation and extrapolation in pre-training. We believe this type of theory may be helpful as an inspiration to explore why in the high diversity regime there seems to be a difference between the performance of meta-learning and transfer learning via pre-trained methods.

Finally, the work by \cite{empericalstudypresentation2020, Miranda2020} first demonstrated that there exist synthetic datasets capable of exhibiting higher degrees of adaptation compared to the original work by \cite{Raghu2020}. Their main focus was on comparing adapted MAML models vs. unadapted MAML models, a difference from our approach in this paper.

Previous work demonstrated that in datasets with low diversity, the difference between meta learning and pre-training is small \citep{curse_low_div}.
While we substantiate these results to a degree, we introduce a nuance that, based on dataset diversity, either technique can outperform the other. 
Finally, we provide the final piece of evidence to complete their story \citep{curse_low_div}, for high diversity datasets.
Crucially, we include the large scale meta-dataset (mds) and demonstrate that merging/unioning datasets is an effective mechanism for increasing the diversity of a dataset. 

BiT \citep{kolesnikov2020bit} is a study that demonstrates good performance on a wide set of datasets (20) using pre-training on a large (JFT-300M) scale vision dataset. 
They fine-tune the entire network (with SGD and momentum) during adaptation and provide heuristics for choosing the hyperparameters with the HyperRule heuristic. 
The main contrast with our work is that they do not employ a model-agnostic approach for pre-training as we do. 
We note they use a large dataset for pre-training, and it's important to use such a dataset for the training of MAML to be able to do a fair comparison between MAML and PT.
Our experiments on meta-dataset suggest that on large-scale diverse dataset MAML might be marginally better than PT. 

Memory efficient meta-learning with large images \citep{bronskill2021memory} demonstrates that if one subsamples the support set (using their method called LIME) to meta-train many meta-learning algorithms, then one can match the performance of a pre-training network that has been fine-tuned with 50 steps. 
The main contrast between their work and ours is:
1. They use confidence intervals to separate pre-training methods vs meta-learning methods, while we use effect size
2. We add another level of structure to the analysis by separating the results in datasets that have a low diversity vs a high diversity.
This analysis shows that meta learning in fact can outperform pre-training, although being small when using the effect size as the measuring metric.
We posit that our work, in conjunction with \cite{bronskill2021memory}, provides a complete perspective on meta-learning --  where we conjecture that meta-learning methods in general marginally outperform pre-training methods. 
Their work \citep{bronskill2021memory} supports our counter-narrative that pre-training methods are always better.

The Vendi Score is a recently proposed diversity score different from the diversity coefficient proposed in \cite{curse_low_div}.
The Vendi score is a more sophisticated aggregation method than an expectation given pair-wise comparisons.
Their aggregation score is interesting, but it is unclear what the advantages of it are compared to a simpler expectation.
For a sample of $n$ already embedded tasks (or data points), the Vendi score takes $O(n^3)$ (due to the use of eigenvalue computation), while ours uses expectation, which takes $O(n^2 - n / 2) = O(n^2)$. 
We hope to explore the Vendi score in future work and compare it with the expectation aggregation score.
However, the main weakness of the Vendi score that previous work address \citep{curse_low_div} is the use of Task2Vec \citep{task2vec} to compute embeddings of tasks. 
The Vendi score assumes one already has such a comparison by assuming a Kernel/Grahm Matrix and circumvents this problem.
Their formulation also implies their analysis is mostly focused on individual data point diversity, while the diversity coefficient also works embedding tasks, batches, and even entire datasets. 


\subsection{Dataset composition details} \label{datasets}
Here we detail the composition of our high diversity datasets, and outline what the dataset acronyms stand for in the subsequent sections.
The method we used was taking the union as in \citep{rfs} of different datasets.
We used global labels \citep{global_labels} during training. 

\begin{itemize}
    \item hdbi stands for High-Diversity Benchmark number $i$.
    \item MIO stands for combining MiniImagenet \citep{miniimagenet} and Omniglot \citep{omniglot}.
    \item MICOD stands for combining MiniImagenet \citep{miniimagenet}, Cifar-fs \citep{cifarfs}, Omniglot \citep{omniglot}, and Delaunay \citep{gontier2022delaunay}.
    \item AFDO stands for combining fgvcAircraft \citep{aircraft}, vggFlower \citep{flower}, Delaunay \citep{gontier2022delaunay}, and Omniglot \citep{omniglot}.
    \item AFTO stands for combining fgvcAircraft \citep{aircraft}, vggFlower \citep{flower}, describableTextures \citep{dtd}, and Omniglot \citep{omniglot}.
    \item CADO stands for combining  Cifar-fs \citep{cifarfs}, FGVCAircraft \citep{aircraft}, Delaunay \citep{gontier2022delaunay}, and Omniglot \citep{omniglot}.
    \item CAVDO stands for combining Cifar-fs \citep{cifarfs}, FGVCAircraft \citep{aircraft}, VGGFlower \citep{flower}, DescribableTextures \citep{dtd}, Omniglot \citep{omniglot}.
    \item MICOVA stands for combining  MiniImagenet \citep{miniimagenet}, Cifar-fs \citep{cifarfs}, Omniglot \citep{omniglot}, VGGFlower \citep{flower}, FGVCAircraft \citep{aircraft}.
    \item MDS stands for Meta-Dataset \citep{Triantafillou2019}.
    \item DTD stands for Describable Textures Dataset \citep{dtd}.
    \item VGGFlower is the alternative name for fgvcFlower \citep{flower}.
    \item VGGAir stands for combining VGGflower \citep{flower} and fgvcAircraft \citep{aircraft}.
    \item VGGDTD stands for combining VGGflower \citep{flower} and DTD \citep{dtd}.
\end{itemize}

\subsubsection{Diversities of datasets}

In this section, we present the diversities computed on the datasets we studied.
We used the Task2Vec diversity coefficient described in Section \ref{background} and \cite{curse_low_div}.
We divide the results into datasets with low diversity in Table \ref{tab:divs_low_div} and high diversity in Table \ref{tab:divs_high_div}.
We do the division of low vs high at $0.146$ Task2Vec diversity (using a pre-trained (pt) Resnet18 (Renset18 pt)), because that was approximately the average diversity.

\begin{table*}[h]
\centering
\caption{
\textbf{Diversity Coefficient of low diversity datasets.} We present results with 95\% confidence intervals.
We use a Resnet18 and Resnet34 pre-trained (pt) on ImageNet as the backbone to calculate the Fisher Information Matrix (FIM) needed for the Task2Vec task embeddings for the Diversity Coefficient.
}
\begin{tabular}{|c|c|c|}
\hline
Dataset & Diversity (Resnet18 pt) & Diversity (Resnet34 pt) \\ \hline
CIFAR-FS & 0.106 $\pm$ 0.00166 & 0.0890 $\pm$ 0.00199 \\
FC100 &  0.107 $\pm$ 0.00149 & 0.0903 $\pm$ 0.00389 \\
mini-ImageNet & 0.119 $\pm$ 0.00213  & 0.102 $\pm$ 0.00163 \\
tiered-ImageNet & 0.124 $\pm$ 0.00219  & 0.105 $\pm$ 0.00161 \\

Aircraft &  0.110 $\pm$ 0.00127 & 0.0932 $\pm$ 0.00109 \\
Flower & 0.138 $\pm$ 0.00288 & 0.117 $\pm$ 0.00234 \\
DTD & 0.129 $\pm$ 0.00227 & 0.111 $\pm$ 0.00228 \\
Delaunay & 0.128 $\pm$ 0.00268 & 0.1078 $\pm$ 0.00196 \\
CuBirds & 0.120 $\pm$ 0.00161 & 0.104 $\pm$ 0.00149 \\

VGGAir & 0.141 $\pm$ 0.00131 & 0.120 $\pm$ 0.00129 \\
VGGDTD & 0.135 $\pm$ 0.00105 & 0.119 $\pm$ 0.00107 \\
\hline
\end{tabular}
\label{tab:divs_low_div}
\end{table*}

\begin{table*}[h]
\centering
\caption{
\textbf{Diversity Coefficient of high diversity datasets.} We present results with 95\% confidence intervals.
We use a Resnet18 and Resnet34 pre-trained (pt) on ImageNet as the backbone to calculate the Fisher Information Matrix (FIM) needed for the Task2Vec task embeddings for the Diversity Coefficient.
}
\begin{tabular}{|c|c|c|}
\hline
Dataset & Diversity (resnet18pt) & Diversity (resnet34pt) \\ \hline
Omniglot & 0.167 $\pm$ 0.00579 & 0.139 $\pm$ 0.00387 \\

MIO & 0.188 $\pm$ 0.00416 & 0.161 $\pm$ 0.00351 \\

hdb4-MICOD &  0.174 $\pm$ 0.00420 & 0.154 $\pm$ 0.00381 \\
hdb6-AFDO &  0.179 $\pm$ 0.00255 & 0.155 $\pm$ 0.00218 \\
hdb7-AFTO & 0.186 $\pm$ 0.00276 & 0.146 $\pm$ 0.00233 \\
hdb8-CADO & 0.173 $\pm$ 0.00278 & 0.153 $\pm$ 0.00236 \\
hdb9-CAVDO & 0.177 $\pm$ 0.00256 & 0.139 $\pm$ 0.00199 \\
hdb10-MICOVA & 0.170 $\pm$ 0.00262 & 0.137 $\pm$ 0.00214 \\

MDS & 0.173 $\pm$ 0.00282 & 0.149 $\pm$ 0.00252 \\
\hline
\end{tabular}
\label{tab:divs_high_div}
\end{table*}

\subsection{Justification for Effect Size and Decision Rule}\label{sec:es_just}

The principal rationale behind our choice for using effect size lies in the sizable sample/batch size we employed, ranging from 300-500. 
When we used t-tests, we obtained p-values and confidence intervals equal to zero.
Therefore, we cannot meaningfully use these statistical methods and over-reject the null hypothesis, a known issue in statistics \citep{pvalue}.
In other words, in studies with large sample sizes, even tiny, unimportant differences can be statistically significant. 
Reporting effect size prevents this type of misleading conclusion \citep{pvalue}.

We also seek a data-centric perspective to our analysis.
Therefore, it was paramount that our analysis was robust across datasets.
Using effect size allows for this comparison because it's a standardized measure, so it's not influenced by 
the units of measurement.

Effect size provides information about the magnitude or strength of the difference or relationship between variables, going beyond the mere existence of an effect. 
It offers more informative insights than solely determining whether a difference is significant.

In the end, every statistical decision test will have an threshold value that needs to be chosen and justified.
For t-tests, it's the p-value, commonly set to 0.05. 
For confidence intervals is the confidence level, commonly set to 95\%.
For effect size, the standard used values are 0.2, 0.5, 0.8 \citep{andrade2020meandifference}.
In this setting, we choose the difference to be the standardized 1 \%.
We choose this value because it is the common performance gained needed in machine learning conferences, these papers are some evidence \citep{rodríguez2020embedding, li2021universal}.
However, note that the choice is arbitrary, and therefore it is important to report all absolute effect sizes and raw meta-test accuracies.
We do this in our work, and the meta-test accuracies are in the supplementary section \ref{sec:exp_details}.
We finalize this section by reminding the reader all statistical tests have assumptions and are imperfect.

\subsection{Experimental Details and Results}\label{sec:exp_details}
In this section we present experimental and hyperparameter details of all the experiments conducted. We present the raw meta-test accuracies and effect sizes obtained, which are used when comparing PT vs MAML models in the main body of the text Section \ref{experiments}.
We split our experiments into 4 major categories: low diversity experiments with first-order MAML (fo-MAML), low diversity experiments with higher-order MAML (ho-MAML), high diversity experiments with varying Resnet architectures, high diversity experiments with varying CNN architectures.

\subsubsection{Low Diversity Experiments with First-Order MAML}\label{sec:res_low_fo}
We used the Resnet12 architecture provided by \cite{rfs}, which has 1,427,525 parameters. 
The Adam optimizer \citep{adam} was utilized with a constant learning rate of 1e-3. 
No learning rate scheduler was used. 
Training was performed for 600,000 iterations for pre-training 
and 160,000 first-order MAML iterations, with a batch size of 256. 
The outer loop consisted of 130,000 MAML iterations. 
We used a meta-batch size of 300 few-shot learning tasks.
We used an inner learning rate of 0.1 and 5 inner steps.
No weight decay was applied. 
Training was performed on a single NVIDIA PU with at most 48GB memory select by a HPC automatically.
All experiments were trained to convergence (less than 0.01 loss) and took on average at most 1 week. 
All implementations were done in PyTorch \citep{pytorch}.

We report meta-test accuracies (with 95\% confidence intervals) of the Resnet model trained with Pre-training (PT) vs. first-order (FO) MAML model on different low diversity datasets in Table \ref{tab:meta_test_acc_low_div_fo_maml}, and report the effect sizes in Table \ref{tab:comparison_fo_maml_es}. 
Overall, the effect size for H0 (no diff.) is 0.029, for H1 (pt) is 0.778, and for H1 (meta) is -0.411. 
However, when averaging all low diversity results (as done in section \ref{experiments_summary}) where there was a difference (H1 pt or meta), the overall effect size is 0.103, favoring pre-training.

\begin{table}[h]
\centering
\caption{
\textbf{Meta-Test accuracy of a Pre-trained (PT) model vs. a first-order (FO) MAML model with 95\% confidence intervals on low diversity few-shot learning vision datasets.}
}
\begin{tabular}{|c|c|c|c|}
\hline
Dataset (Model) & PT (test acc.) & MAML5 (test acc.) & MAML10 (test acc.) \\ \hline
CIFAR-FS (Resnet12) & 0.755 $\pm$ 0.0102 & 0.779 $\pm$ 0.00975 & 0.786 $\pm$ 0.00996 \\
FC100 (Resnet12) & 0.438 $\pm$ 0.00949 & 0.458 $\pm$ 0.00931 & 0.459 $\pm$ 0.00988 \\
mini-ImageNet (Resnet12) & 0.719 $\pm$ 0.00893 & 0.685 $\pm$ 0.00947 & 0.706 $\pm$ 0.0104 \\
tiered-ImageNet (Resnet12) & 0.788 $\pm$ 0.00945 & 0.769 $\pm$ 0.0107 & 0.786 $\pm$ 0.0107 \\

Aircraft (Resnet12) & 0.592 $\pm$ 0.010 & 0.659 $\pm$ 0.013 & 0.685 $\pm$ 0.011 \\
Flower (Resnet12) & 0.928 $\pm$ 0.005 & 0.856 $\pm$ 0.008 & 0.870 $\pm$ 0.007 \\
DTD (Resnet12) & 0.610 $\pm$ 0.011 & 0.511 $\pm$ 0.012 & 0.528 $\pm$ 0.011 \\
Delaunay (Resnet12) & 0.735 $\pm$ 0.010 & 0.614 $\pm$ 0.012 & 0.632 $\pm$ 0.010 \\
CuBirds (Resnet12) & 0.787 $\pm$ 0.008 & 0.829 $\pm$ 0.008 & 0.821 $\pm$ 0.009 \\

VGGAir (Resnet12) & 0.727 $\pm$ 0.027 & 0.745 $\pm$ 0.019 & 0.760 $\pm$ 0.019 \\
VGGDTD (Resnet12) & 0.737 $\pm$ 0.019 & 0.701 $\pm$ 0.022 & 0.701 $\pm$ 0.021 \\
\hline
\end{tabular}
\label{tab:meta_test_acc_low_div_fo_maml}
\end{table}

\begin{table*}[h!]
\centering
\caption{
\textbf{Effect size between a Pre-trained (PT) model vs. first-order (FO) MAML model on low diversity few-shot learning vision datasets.}
Here the null hypothesis (H0) refer to the cases when both meta learning and pre-training perform equivalently according to our decision rule. H1 (pt) and H1 (meta) refer to the cases where pre-training and meta learning performed better respectively.
}
\begin{tabular}{|c|c|c|}
\hline
Dataset (model) & ES (PT vs MAML5) & ES (PT vs MAML10) \\ \hline

CIFAR-FS (Resnet12) & -0.266 (H1 meta) & -0.342 (H1 meta) \\
FC100 (Resnet12) & -0.251 (H1 meta) & -0.248 (H1 meta) \\
mini-ImageNet (Resnet12) & 0.413 (H1 pt) & 0.149 (H1 pt) \\
tiered-ImageNet (Resnet12) & 0.218 (H1 pt) & 0.0290 (H0 no diff.) \\

Aircraft (Resnet12) & -0.671 (H1 meta) & -1.014 (H1 meta) \\
Flower (Resnet12) & 1.224 (H1 pt) & 1.125 (H1 pt) \\
DTD (Resnet12) & 1.332 (H1 pt) & 1.147 (H1 pt) \\
Delaunay (Resnet12) & 1.290 (H1 pt) & 1.262 (H1 pt) \\
CuBirds (Resnet12) & -0.572 (H1 meta) & -0.452 (H1 meta) \\

VGGAir (Resnet12) &  -0.105 (H1 meta) & -0.195 (H1 meta) \\
VGGDTD (Resnet12) & 0.200 (H1 pt) & 0.203 (H1 pt) \\
\hline
\end{tabular}
\label{tab:comparison_fo_maml_es}
\end{table*}

\subsubsection{Low Diversity Experiments with Higher-Order MAML}
We use the same architecture and hyperparameters used for the fo-MAML experiments as in Section \ref{sec:res_low_fo}. We report meta-test accuracies (with 95\% confidence intervals) of the Resnet model trained with Pre-training (PT) vs. higher-order (HO) MAML model on different low diversity datasets in Table \ref{tab:meta_test_acc_low_div_ho_maml}, and report the effect sizes in Table \ref{tab:comparison_ho_maml_es}. 
Overall, no experiment resulted in a decision of H0. The overall effect effect size for H1 (pt) is 0.669, and for H1 (meta) is -0.717. 
However, when averaging all low diversity results (as done in Section \ref{experiments_summary}) where there was a difference (H1 pt or meta), the overall effect size is 0.103, favoring pre-training.

\begin{table}[h]
\centering
\caption{
\textbf{Meta-Test accuracy of a Pre-trained (PT) model vs. a higher-order (HO) MAML model with 95\% confidence intervals on low diversity few-shot learning vision datasets.}
}
\begin{tabular}{|c|c|c|c|c|}
\hline
Dataset (Model) & PT (test acc.) & MAML5 (test acc.) & MAML10 (test acc.) \\ \hline
CIFAR-FS (Resnet12) & 0.753 $\pm$ 0.00941 & 0.804 $\pm$ 0.00982 & 0.809 $\pm$ 0.0107 \\
FC100 (Resnet12)  & 0.432 $\pm$ 0.0102 & 0.503 $\pm$ 0.0100 & 0.489 $\pm$ 0.00988 \\
mini-ImageNet (Resnet12) & 0.721 $\pm$ 0.00889 & 0.704 $\pm$ 0.0100 & 0.732 $\pm$ 0.00952 \\
tiered-ImageNet (Resnet12) & 0.791 $\pm$ 0.00922 & 0.771 $\pm$ 0.0103 & 0.695 $\pm$ 0.0178 \\

Aircraft (Resnet12) & 0.576 $\pm$ 0.0116 & 0.647 $\pm$ 0.0127 & 0.667 $\pm$ 0.0112 \\
Flower (Resnet12) & 0.921 $\pm$ 0.00534 & 0.902 $\pm$ 0.00597 & 0.899 $\pm$ 0.00581 \\
DTD (Resnet12) & 0.600  $\pm$ 0.0156 & 0.501  $\pm$ 0.0162 & 0.519  $\pm$ 0.0159 \\
Delaunay (Resnet12) & 0.734  $\pm$ 0.00984 & 0.655  $\pm$ 0.00981 & 0.665 $\pm$ 0.00986 \\
CuBirds (Resnet12) & 0.785 $\pm$ 0.00839 & 0.857 $\pm$ 0.00721 & 0.857 $\pm$ 0.00726 \\
\hline
\end{tabular}
\label{tab:meta_test_acc_low_div_ho_maml}
\end{table}

\begin{table*}[h!]
\centering
\caption{
\textbf{Effect size between a Pre-trained (PT) model vs higher-order (HO) MAML model on low diversity few-shot learning vision datasets.}
Here the null hypothesis (H0) refer to the cases when both meta learning and pre-training perform equivalently according to our decision rule. H1 (pt) and H1 (meta) refer to the cases where pre-training and meta learning performed better respectively.
}
\begin{tabular}{|c|c|c|c|c|}
\hline
Dataset (model) & ES (PT vs MAML5) & ES (PT vs MAML10)\\ \hline
CIFAR-FS (Resnet12) & -0.602 (H1 meta) & -0.628 (H1 meta) \\
FC100 (Resnet12)  & -0.800 (H1 meta) & -0.643 (H1 meta) \\
mini-ImageNet (Resnet12) & 0.205 (H1 pt) & -0.126 (H1 meta) \\
tiered-ImageNet (Resnet12) & 0.236 (H1 pt) & 0.768 (H1 pt) \\

Aircraft (Resnet12) & -0.667 (H1 meta) & -0.908 (H1 meta) \\ 
Flower (Resnet12) & 0.382 (H1 pt) & 0.465 (H1 pt) \\
DTD (Resnet12) & 1.240 (H1 pt) & 1.020 (H1 pt) \\
Delaunay (Resnet12) & 0.912 (H1 pt) & 0.793 (H1 pt) \\
CuBirds (Resnet12) & -1.043 (H1 meta) & -1.044 (H1 meta) \\
\hline
\end{tabular}
\label{tab:comparison_ho_maml_es}
\end{table*}

\subsubsection{High Diversity Experiments with ResNet}
For experiments with ResNet12, we utilized the ResNet12 architecture from \citet{rfs}, which has 1,427,525 parameters.
The Adam optimizer \citep{adam} was used with a learning rate of 1e-3 without any learning rate decay.
For pre-training, we trained for 1 million iterations with a batch size of 256. 
Unless otherwise stated, we used first-order MAML \citep{maml}, for which we trained for 300,000 iterations also with a batch size of 256.
The MAML outer loop consisted of 5 inner update steps with an inner learning rate of 0.1.
We used a meta-batch size 300 few-shot learning tasks.
No weight decay was applied.
We annealed the learning rate with a cosine scheduler with scheduler freq 2000 with minimum learning rate 1e-5 (similar to MAML++).
All models were trained to convergence on a single NVIDIA GPU with at least 48GB of memory allocated by the cluster scheduler.
Training took approximately 1-2 week to converge for both pre-training and MAML.
All implementations were done in PyTorch \citep{pytorch}.

We used ResNet50 for our experiments with MDS, where we utilized the ResNet50 architecture from \citet{rfs} which has 50,685,637 parameters.
The Adafactor optimizer \citep{adafactor} was used with default settings and no learning rate decay.
We used Adafactor from FairSeq because it had a setting with no hyperparameter choices, the memory benefits that we needed given our compute and the evidence of previous work showing the training was 2.5-fold faster \citep{miranda2023transformer}.
For pre-training, we trained for 300,000 iterations with a batch size of 256.
Unless otherwise stated, we used first-order MAML \citep{maml}, for which we trained for 140,000 iterations also with a batch size of 256.
The MAML outer loop consisted of 5 inner update steps with Adafactor's default inner learning rate.
We used a meta-batch size 300 few-shot learning tasks.
We used Adafactor default annealing scheduler in FairSeq.
Due to computational constraints, we limited the number of random seeds to 4 -- especially given that MDS combined 10 large scale vision datasets that includes ImageNet.
Pre-training and MAML training took approximately 1 month each to converge on NVIDIA GPUs with 48GB memory allocated automatically by the cluster scheduler.
All implementations were done in PyTorch \citep{pytorch}.

Meta-test accuracies (with 95\% confidence intervals) of a Pre-training (PT) vs. MAML Resnet model on different high diversity datasets are in Table \ref{tab:meta_test_acc_high_div}, and effect sizes are reported in Table \ref{tab:comparison_high_div}. 
For several rows we report averaged effect sizes over multiple seeds. When running pt for $p$ seeds and maml for $m$ seeds we report averages over all possible $p*m$ comparisons.
Overall, the effect size for H0 (no diff.) is 0.0721, for H1 (pt) is 0.0638, and for H1 (meta) is -0.155. 
However, when averaging all high diversity results (as done in Section \ref{experiments_summary}) where there was a difference (H1 pt or meta), the overall effect size is -0.107, favoring meta.

\begin{table}[h]
\centering
\caption{
\textbf{Meta-Test accuracy of a Pre-trained (PT) model vs. a MAML model with 95\% confidence intervals on high diversity few-shot learning vision datasets.}
}
\begin{tabular}{|c|c|c|c|c|}
\hline
Dataset (Model) & PT (test acc.) & MAML5 (test acc.) & MAML10 (test acc.) \\ 
\hline
Omniglot (Resnet12) & 0.993 $\pm$ 0.00148 & 0.993 $\pm$ 0.00139 & 0.992 $\pm$ 0.00164 \\
Omniglot (Resnet12, ho maml) & 0.994 $\pm$ 0.00110 & 0.985 $\pm$ 0.00219 & 0.988 $\pm$ 0.00180 \\

MIO (Resnet12)&  0.845 $\pm$ 0.0121 & 0.849 $\pm$ 0.0136 & 0.848 $\pm$ 0.0133 \\

hdb4-MICOD (Resnet12) & 0.778 $\pm$ 0.0124 & 0.781 $\pm$ 0.0124 & 0.786 $\pm$ 0.0119 \\ 

hdb6-AFDO (Resnet12) & 0.786 $\pm$ 0.0205 & 0.802 $\pm$ 0.0190 & 0.782 $\pm$ 0.0178 \\
hdb7-AFTO (Resnet12)& 0.756 $\pm$ 0.0227 & 0.745 $\pm$  0.0226 & 0.779 $\pm$ 0.0216 \\
hdb8-CADO (Resnet12) & 0.711 $\pm$ 0.0218 & 0.744 $\pm$ 0.0227 & 0.733 $\pm$ 0.0208 \\
hdb9-CAVDO (Resnet12) & 0.772 $\pm$ 0.0210 & 0.771 $\pm$ 0.0207 & 0.762 $\pm$ 0.0211 \\
hdb10-MICOVA (Resnet12) & 0.713 $\pm$ 0.0244 & 0.764 $\pm$ 0.0177 & 0.766 $\pm$ 0.0167 \\

MDS (Resnet50, seed 1) & 0.775 $\pm$ 0.0133 & 0.762 $\pm$ 0.0133  & 0.768 $\pm$ 0.0144 \\
MDS (Resnet50, seed 2) & 0.752 $\pm$ 0.0138 & 0.758 $\pm$ 0.0150 & 0.768 $\pm$ 0.0144 \\
MDS (Resnet50, seed 3) & 0.750 $\pm$ 0.0141 & 0.759 $\pm$ 0.0151 & 0.772 $\pm$ 0.0152  \\
MDS (Resnet50, seed 4) & 0.765 $\pm$ 0.0135 & 0.762 $\pm$ 0.0147 & 0.776 $\pm$ 0.0143  \\
\hline
\end{tabular}
\label{tab:meta_test_acc_high_div}
\end{table}

\begin{table*}[h!]
\centering
\caption{
\textbf{
Effect size between a Pre-trained (PT) model vs MAML model on high diversity few-shot learning vision datasets.} 
Here the null hypothesis (H0) refer to the cases when both meta learning and pre-training perform equivalently according to our decision rule. H1 (pt) and H1 (meta) refer to the cases where pre-training and meta learning performed better respectively.
}
\begin{tabular}{|c|c|c|}
\hline
Dataset (model) & ES (PT vs MAML5) & ES (PT vs MAML10) \\ \hline

Omniglot (Resnet12) & 0.00702 (H0 no diff.) & 0.0679 (H0 no diff.) \\
Omniglot (Resnet12, ho maml) & 0.577 (H0 no diff.) & 0.468 (H0 no diff.) \\

MIO (Resnet12) & -0.0197 (H0 no diff.) & -0.0161 (H0 no diff.) \\

hdb4-MICOD (Resnet12) & -0.0166 (H0 no diff.) & -0.0559 (H0 no diff.) \\

hdb6-AFDO (Resnet12) & -0.0919 (H1 meta) & 0.0242 (H0 no diff.) \\
hdb7-AFTO (Resnet12) & 0.0528 (H1 pt) & -0.121 (H1 meta) \\
hdb8-CADO (Resnet12) & -0.167 (H1 meta) & -0.116 (H1 meta) \\
hdb9-CAVDO (Resnet12) & 0.00798 (H0 no diff.) & 0.0552 (H1 pt) \\
hdb10-MICOVA (Resnet12) & -0.287 (H1 meta) & -0.308 (H1 meta) \\

hdb12 (Resnet12, 9 seeds) & -0.145 (H1 meta) & -0.190 (H1 meta) \\
hdb13 (Resnet12, 9 seeds) & -0.02955 (H0 no diff.) & -0.0345 (H0 no diff.) \\
hdb14 (ResNet12, 9 seeds) & 0.0371 (H0 no diff.) & -0.05745 (H0 no diff.) \\
MDS (Resnet50, 4 seeds) & 0.001375 (H0 no diff.) & -0.064275 (H1 meta) \\
\hline
\end{tabular}
\label{tab:comparison_high_div}
\end{table*}

\subsubsection{High Diversity Experiments with CNNs}
We utilized the 5CNN architecture proposed in \cite{rfs}, which consists of CNNs with 5 convolutional layers, with varying filter sizes.
The Adam optimizer \cite{adam} was used with a learning rate of 1e-3 without any learning rate decay. 
A batch size of 256 was used for both pre-training and MAML training. 
No weight decay was applied. 
For pre-training, we trained for 200,000 iterations. 
We used first-order MAML, for which we trained for 100,000 iterations with an inner loop of 5 steps and an inner learning rate of 0.1. 
We used a meta-batch size of 500 few-shot learning tasks.
We annealed the learning rate with a cosine scheduler with scheduler freq 2000 with minimum learning rate 1e-5 (similar to MAML++).
All models were trained to convergence, which took approximately 1 week on a single NVIDIA GPU with at least 48GB of memory allocated by the HPC scheduler. 
All implementations were done in PyTorch \cite{pytorch}.

Meta-test accuracies (with 95\% confidence intervals) of 5CNNs trained with Pre-training (PT) vs. MAML on the MICOD high diversity dataset are in Table \ref{tab:comparison_5cnn_micod_test_acc}. We report the effect sizes in Table \ref{tab:comparison_5cnn_micod_es}. 
Overall, the effect size for H0 (no diff.) is 0.00922, for H1 (pt) is 0.0795, and for H1 (meta) is -0.192. 
However, when averaging all high diversity results (as done in Section \ref{experiments_summary}) where there was a difference (H1 pt or meta), the overall effect size is -0.107, favoring meta learning.

\begin{table}[h]
\centering
\caption{
\textbf{Meta-Test accuracy of a Pre-trained (PT) model vs. a MAML model with 95\% confidence intervals on the high diversity MICOD few-shot learning vision dataset.}
}
\begin{tabular}{|c|c|c|c|}
\hline
Filter Size (Dataset) & PT (test acc.) & MAML5 (test acc.) & MAML10 (test acc.) \\ \hline
2 (micod) & 0.481 $\pm$ 0.0205 & 0.493 $\pm$ 0.0197  & 0.467 $\pm$ 0.0184 \\
6 (micod) & 0.588 $\pm$ 0.0169 & 0.626 $\pm$ 0.0189  & 0.608  $\pm$ 0.0178 \\
8 (micod) & 0.606 $\pm$ 0.0161 & 0.591 $\pm$ 0.0178 & 0.607 $\pm$ 0.0184 \\
16 (micod) & 0.655 $\pm$ 0.0149 & 0.678 $\pm$ 0.0154 & 0.681 $\pm$ 0.0157 \\
32 (micod) & 0.689 $\pm$ 0.0151 & 0.682 $\pm$ 0.0150 & 0.701 $\pm$ 0.0154 \\
64 (micod) & 0.694 $\pm$ 0.0135 & 0.704 $\pm$ 0.0155 & 0.718 $\pm$ 0.0152 \\
256 (micod) & 0.711 $\pm$ 0.0139 & 0.702 $\pm$ 0.0163 & 0.695 $\pm$ 0.0156 \\
512 (micod) & 0.653 $\pm$ 0.0175 & 0.718 $\pm$ 0.0158 & 0.724 $\pm$ 0.0154 \\
\hline
\end{tabular}
\label{tab:comparison_5cnn_micod_test_acc}
\end{table}

\begin{table*}[h!]
\centering
\caption{
\textbf{Effect size between a Pre-trained (PT) model vs MAML model on the high diversity MICOD few-shot learning vision dataset.}
Here the null hypothesis (H0) refer to the cases when both meta learning and pre-training perform equivalently according to our decision rule. H1 (pt) and H1 (meta) refer to the cases where pre-training and meta learning performed better respectively.
}
\begin{tabular}{|c|c|c|}
\hline
Filter Size (Dataset) & ES (PT vs MAML5) & ES (PT vs MAML10) \\ \hline
2 (micod) & -0.0533 (H1 meta) & 0.0624 (H1 pt)  \\
6 (micod) & -0.184 (H1 meta) & -0.100 (H1 meta) \\
8 (micod) & 0.0794 (H1 pt) & -0.00121 (H0 no diff) \\
16 (micod) & -0.131 (H1 meta) & -0.149 (H1 meta) \\
32 (micod)  & 0.0401 (H0 no diff) & -0.0689 (H1 meta) \\
64 (micod) & -0.0588 (H0 no diff) & -0.145 (H1 meta) \\
256 (micod) & 0.0568 (H0 no diff) & 0.0969 (H1 pt) \\
512 (micod) & -0.341 (H1 meta) & -0.376 (H1 meta) \\
\hline
\end{tabular}
\label{tab:comparison_5cnn_micod_es}
\end{table*}

\subsection{Investigations on GPT-2}\label{sec:gpt}
Large language models (LLMs) including GPT-2 are typically trained in a supervised learning setting where the model is trained to predict the token coming after each token in a sentence\citep{gpt2}. 
This does not naturally translate to $k$-shot learning tasks that most of this paper focusses on. 
In particular, we note that the vocabulary size for each token is in the order of tens of thousands (50257 for GPT-2) which is a lot bigger than the few thousand (1623 for Omniglot \citep{omniglot}) classes MAML is typically used for. Additionally, each token does not have an equal number of instances in the language. Finally and most importantly, training a model to chose between a few classes given example occurrences of those classes, and comparing it against a model trained to predict one class out of 50257 is an apples to oranges comparison.

We however note the primary motivation of the paper to establish the performance of "learning a task" against "learning to learn the same task". Hence instead of using examples of specific target classes as the support set as is usually done for MAML, we use example sequences as the support set and example sequences as the query set. An initial idea is for each batch size of size $b$, we can train the model to learn from the first $b/2$ examples and predict the next $b/2$ correctly. However, since the batches are independent, we don't give the model a chance to learn from context and this implementation of MAML reduces to a harder to optimize version of the usual supervised learning setting.

We solve this issue by separating the support and query sets within examples in a batch instead of within batches. That is, if the token size of the model is $t$, we split each example of $t$ tokens into support and query sets. We make the first $t/2$ tokens of each example as the support set and the next $t/2$ tokens as the query set. Hence given a token size $t$, we train the model to learn to predict the last $t/2$ tokens based on the first $t/2$ tokens. Formally, for each example in the batch, we perform an inner loop optimization on the cross-entropy loss of next-token predictions for the first $t/2$ tokens. We then perform the outer loop optimization on the cross-entropy loss of next-token predictions using the obtained model on the last $t/2$ tokens. 

Evaluation comparison between the two training techniques is done by following a similar approach of inner loop optimization using the first half tokens and reporting accuracy values on the second half post inner loop optimization.

\subsubsection{Experiments}
We compared our meta-learning algorithm as described above against the standard pre-training + head fine-tuning approach (PT) employed for language models. We derived our architecture from NanoGPT\footnote{https://github.com/karpathy/NanoGPT}. 
We set a block size of 32 and created a model with 4 layers and 4 heads per layer with an embedding size of 192 to have a smaller model that is easy for experimentation that we dub mini-GPT2. 
The Adafactor optimizer and scheduler \citep{adafactor} were used with default settings (that is, no hyperparameters were specified) and no learning rate decay. 
In addition due to the memory benefits that we needed given our compute and the evidence of previous work showing the training was 2.5-fold faster \citep{miranda2023transformer}.
Training was performed till visual convergence of loss curves was reached for all experiments. 
For MAML training, we set inner loop learning rate as 1e-3 and performed 5 inner update steps. Training was performed in a distributed setting on 4 NVIDIA PU with at most 48GB memory select by a HPC automatically. 
All implementations were done in PyTorch \citep{pytorch}.

Averaged over 30 seeds, we obtained an effect size of 0.00681 comparing PT with MAML5 and an effect size of -0.0114 comparing PT with MAML10. Both of these effect sizes result in a decision of H0, favoring neither algorithm. We would like to note that the dataset used, openwebtext, has a diversity coefficient of $0.222\pm 0.001$, and is hence classified as a high diversity dataset in our paper.


\subsection{Further testing of the Diversity Coefficient} \label{div_coeff}

\begin{figure*}[h!]
\centering
\includegraphics[width=1.0\linewidth]{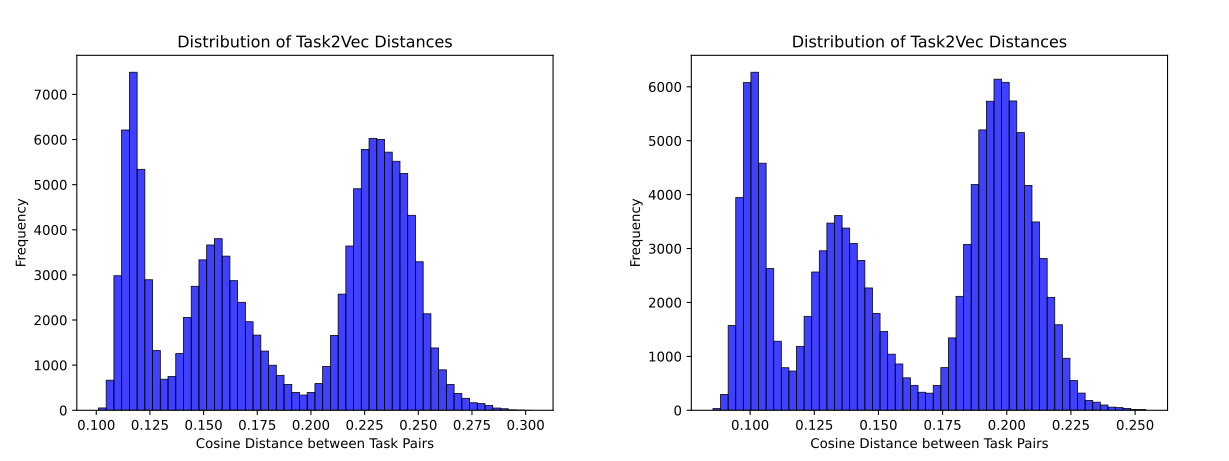}
\caption{ 
\textbf{The Task2Vec distances the histogram cluster in a way that reflect the semantic information of the union of the MiniImagenet and Omniglot (MIO) training datasets.}
Left plot show the histogram of the cosine distance between Task2Vec embeddings made using a Resnet18 backbone pre-trained on Imagenet.
Right show the same, but when using Resnet34. 
The meta-batch size was 500 meaning we used 500 tasks to compute these histograms.
The diversity coefficients are 0.188 $\pm$ 0.00416, 0.161 $\pm$ 0.00351 for the LHS and RHS plots.
}
\label{fig:mio_hist}
\end{figure*}

In this section, we further test if the Task2Vec task embeddings distances cluster in a semantically meaningful way in our dataset MIO.
This test is important because if the Task2Vec embeddings used to compute the diversity coefficient have the structure we'd expect, then it makes the diversity coefficient itself more trustworthy.
The MIO dataset was created by combining the MiniImagenet Omniglot.
Therefore, if Task2Vec is a valid embedding for tasks, we would expect three modes for our histogram:
1. One mode for the distances between tasks generated from MiniImagenet,
2. another mode for distances between tasks generated from Omniglot, 
3. and the last mode for distances between tasks generated from MiniImagenet and tasks generated from Omniglot. 
That is indeed what is seen as shown in Figure \ref{fig:mio_hist}.

One interesting observation is that the average distance between Task2Vec embeddings (i.e. the diversity coefficient) is larger for smaller networks.

\subsection{Discussion (Cont.)}
We'd also like to remark the trade-offs between pre-training and meta-learning that \cite{bronskill2021memory} articulates clearly -- especially given the evidence we present countering the prevailing narrative that advocates for pre-training (PT) methods.
The selection between pre-training vs meta learning strategies should be guided by: the available data, computational resources, and the application's specific requirements. 
For singular task types with ample data and no computational or temporal constraints, fine-tuning within pre-training may suffice. 
Conversely, meta-learning would be more appropriate in scenarios requiring the acquisition of diverse tasks with sparse data on resource-limited devices, or in continual or online learning environments.
In addition, we provide a novel perspective where we show diverse datasets are a scenario when meta-learning methods are marginally better than pre-training methods. 

A potential drawback of our work could be our focus on mainly comparing pre-training (PT) against MAML, instead of considering a wider set of meta-learning algorithms. 
Our justification is as follows:
The current narrative \citep{rfs, Chen2019, Chen, Dhillon2019, Huang2019} implies that PT can beat \textit{any} meta-learning algorithm. 
We'd like to emphasize the word \textit{any}, because it implicates a ``for all" quantifier. 
Therefore, to counter the current narrative, we only need to provide evidence against it (and thus show it is likely false) by considering a \textit{single} meta-learning algorithm.
Therefore, if \textit{PT cannot even beat MAML} -- the simplest of meta-learning algorithms -- it's good evidence against the current narrative. 
Therefore, we only need a \textit{single} meta-learning algorithm to support our conclusions. 
In addition, memory efficient meta-learning \citep{bronskill2021memory} demonstrated that other meta-learning algorithms can match pre-trained models. 
However, as we explained in the related work section, our contributions are novel, complementary and different from \cite{bronskill2021memory} does because:
1. We contextualize our claims in a data-centric framework using diversities over an extensive set of diverse datasets, 
2. Our analysis goes beyond using confidence intervals and reports the effect size, a method we justify in Section \ref{methods}, and,
3. Our novel analysis demonstrates that meta-learning (via MAML) and transfer learning (via PT) can be separated in performance when considering the diversity of the dataset.
In addition, it is reasonable to expect, given the memory efficient results \citep{bronskill2021memory}, that when using their memory efficient methods that a similar trend with other meta-learning algorithms would be observed -- especially given we already showed an initial separation between PT and MAML.
Furthermore, we conduct extensive experiments over a large set of diverse datasets.

Another potential drawback of our work could be the use of the arbitrary 1\% thresholds for our decision rule in Section \ref{methods}.
In machine learning, it is not uncommon to accept papers due to 1\% differences. 
We cite this ICCV 2021 paper \citep{li2021universal} which gives the performance variance of common models on meta-dataset in table 1, which commonly ranges from $0.5$ to $1.0$.
We also cite this ECCV 2020 paper \citep{rodríguez2020embedding} that provides the variance for MiniImageNet, where the standard deviations range around $\approx 0.8$.
However, we'd like to underscore that we \textbf{do not} rely solely on this 1\% cutoff to interpret our experiments.
We also report the raw effect size (and test accuracy) and use the classically accepted ranges for what is considered small effect size \citep{andrade2020meandifference}.
However, there is no silver bullet for statistical analysis. 
All of them have assumptions (e.g. CIs, effect size are best for normally distributed data), and some notion of arbitrary values (e.g. p-values, 95\%-confidence intervals, effect size ranges, our 1\% threshold, etc.) are always chosen to give meaning to the results.
However, one can avoid confirmation bias by choosing the statistical method before the analysis of the experiments is done -- which we do. 
In addition, our main rationale to chose effect size is that one can't manipulate (deliberately or accidentally) the sample size to have the decision rule match our preconceived assumptions -- unlike the p-value in t-tests or confidence intervals where it has been an issue noted here \citep{pvalue} in the large sample size regime. 
Therefore, we attempted to protected our interpretations from confirmation bias.

Another criticism of our work could be the lack of theoretical analysis. 
One reason we choose not to do theoretical analysis is that it is often difficult to give non-vacuous bounds in theory. 
Though some progress has been done here \citep{Chen2021} but does not aim to separate pre-training methods vs meta-learning methods.
However, our experiments have good theoretical motivation inspired from \cite{global_labels}.

The challenges in vectorization of MAML and meta-learning algorithms in general stems from the arises because of the task are different across a meta-batch, so the support set has different arbitrary labels across tasks.
Therefore, vectorization is not straightforward without custom CUDA implementations.
However, instead of vectorizing, one could use the memory-efficient meta-learning strategy \citep{bronskill2021memory} to speed up MAML. 
This is an argument in favor of meta-learning given this new possible memory optimization. 
We leave this promising direction for future work but conjecture this will make meta learning more competitive against transfer learning via PT.


An additional benefit of using the effect size is that it also protects the researchers from confirmation bias.
For example, the researcher cannot deliberately choose a sample/batch size to fit pre-conceived assumptions. 

\subsubsection{Why and when does diversity matter?}

We conjecture two main reasons why diversity matters and explain our rationale:

\begin{enumerate}
    \item \textbf{Conjecture 1: Diversity matters because it enables learning-to-learn.}
    This is the main conjecture we provide evidence for in this paper. 
    The main argument is that if there is high diversity, it means there are diverse tasks in the dataset. 
    For the model to do well, it has to do well on all tasks.
    One way to do it is by learning-to-learn and therefore transfer via PT when challenged with solving a new task. 
    An alternative would be memorizing all the tasks. 
    \item \textbf{Conjecture 2: Diversity matters because it increases chances that training set covers test set.}
    Diversity is a formalization of coverage -- it aims to be the effective (average) number of tasks in a dataset.
    Therefore, the higher the diversity, the more tasks a dataset has.
    This (might) increase the probability that the training set covers the test set and improves performance.
    This exploration of this conjecture is left for future work. 
\end{enumerate}

\subsection{Fair comparison}

Unlike previous, we ensure fair comparison between transfer learning via PT vs meta-learning by comparing only fundamental and model-agnostic algorithms, using a consistent neural network architecture, optimizer, and all models trained to convergence. 

\textbf{Architecture:}
We only compared pre-training vs MAML when they \textbf{both} had the same architecture, and did not use algorithms that require specific architectures during training.
When we used a ResNet we used the one described in \cite{rfs}.

\textbf{Optimization:}
We only compared pre-training vs MAML when they \textbf{both} had the same optimization and scheduling rate.
We used the Adam optimizer for all experiments except for GPT2 and Resnet50 on Meta-DataSet (MDS) where we used Adafactor with default hyperparameters.
We did this because Adafactor has a setting in the Fairseq that requires no hyperparameter search and since Meta-DataSet is a large we scale dataset.
It took us about 1 month to train on MDS with Resnet50.
In addition, previous work demonstrated Adafactor can be fast 1 order of magnitude faster (speedup of ~2 hours to ~39 hours) than Adam with hyperparameter search when training transformer models \citep{miranda2023transformer}.

\begin{figure*}[h!]
\centering
\includegraphics[width=1.0\linewidth]{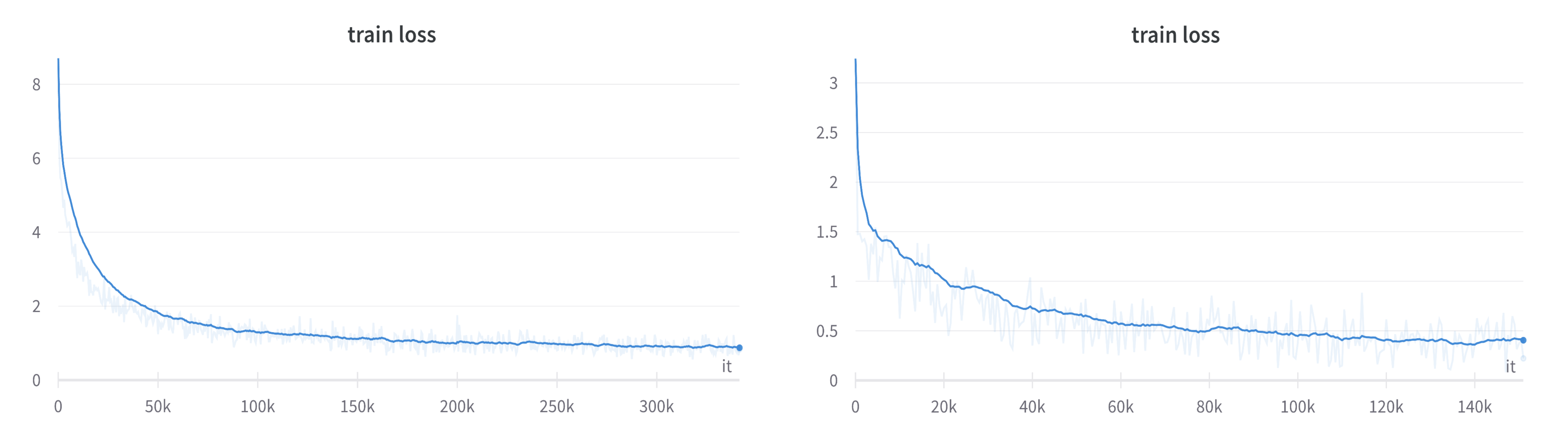}
\caption{
Plot showing convergence of Resnet50 on a high-diversity benchmark Meta-DataSet (MDS) \citep{Triantafillou2019}. 
The left-most plot depicts the training loss curve for the pre-training algorithm, and the rightmost plot depicts the training loss curve for MAML.
}
\label{mds_learning_curve_resnet50}
\end{figure*}

\begin{figure*}[h!]
\centering
\includegraphics[width=1.0\linewidth]{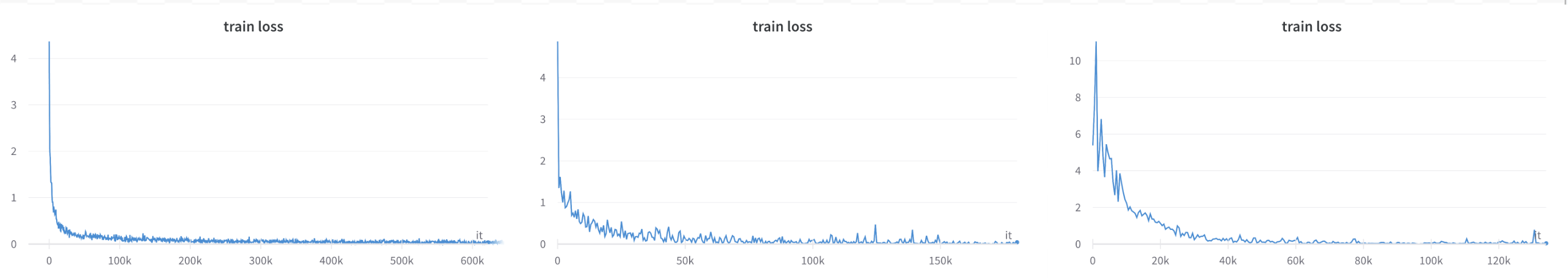}
\caption{
Plot showing convergence of Resnet12 on a low-diversity benchmark (fc100). The left-most plot depicts the training loss curve for the pre-training algorithm, the center plot depicts the training loss curve for first-order MAML, and the rightmost plot depicts the training loss curve for higher-order MAML.
}
\label{res12_fc100_learning_curve_pt_maml}
\end{figure*}

\begin{figure*}[h!]
\centering
\includegraphics[width=1.0\linewidth]{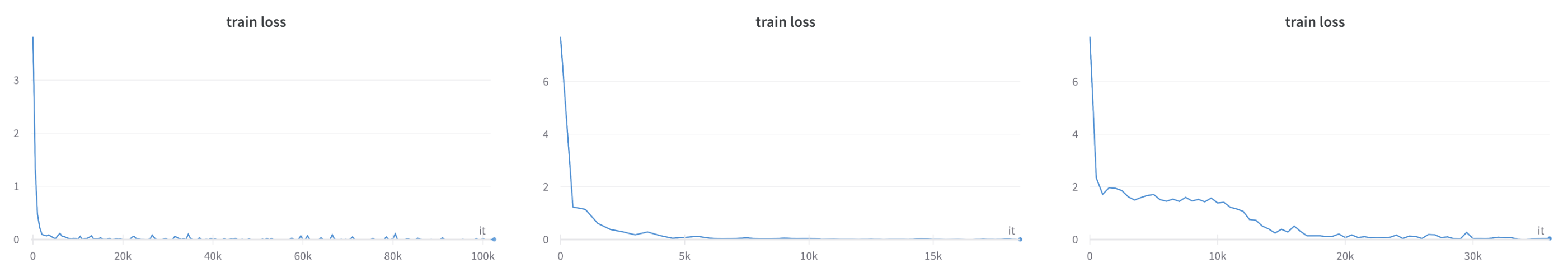}
\caption{
Plot showing convergence of Resnet12 on a low-diversity benchmark (aircraft). The left-most plot depicts the training loss curve for the pre-training algorithm, the center plot depicts the training loss curve for first-order MAML, and the rightmost plot depicts the training loss curve for higher-order MAML.
}
\label{res12_aircraft_learning_curve_pt_maml}
\end{figure*}

\begin{figure*}[h!]
\centering
\includegraphics[width=1.0\linewidth]{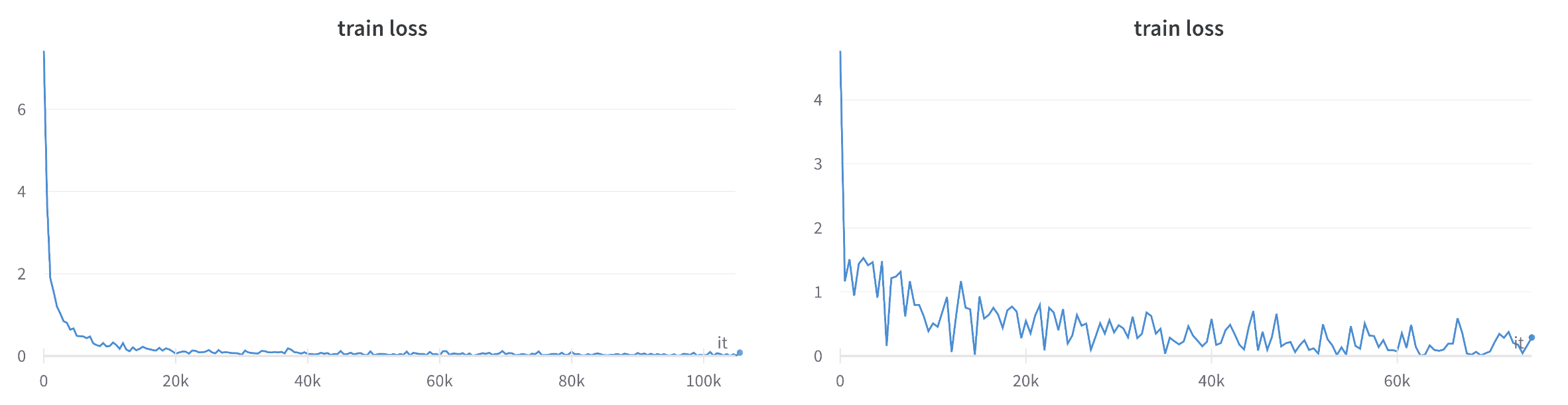}
\caption{
Plot showing convergence of Resnet12 on a high-diversity benchmark (hdb8-cado).The left plot depicts the training loss curve for the pre-training algorithm and the right plot depicts the training loss curve for MAML.
}
\label{res12_hdb8_learning_curve_pt_maml}
\end{figure*}

\begin{figure*}[h!]
\centering
\includegraphics[width=1.0\linewidth]{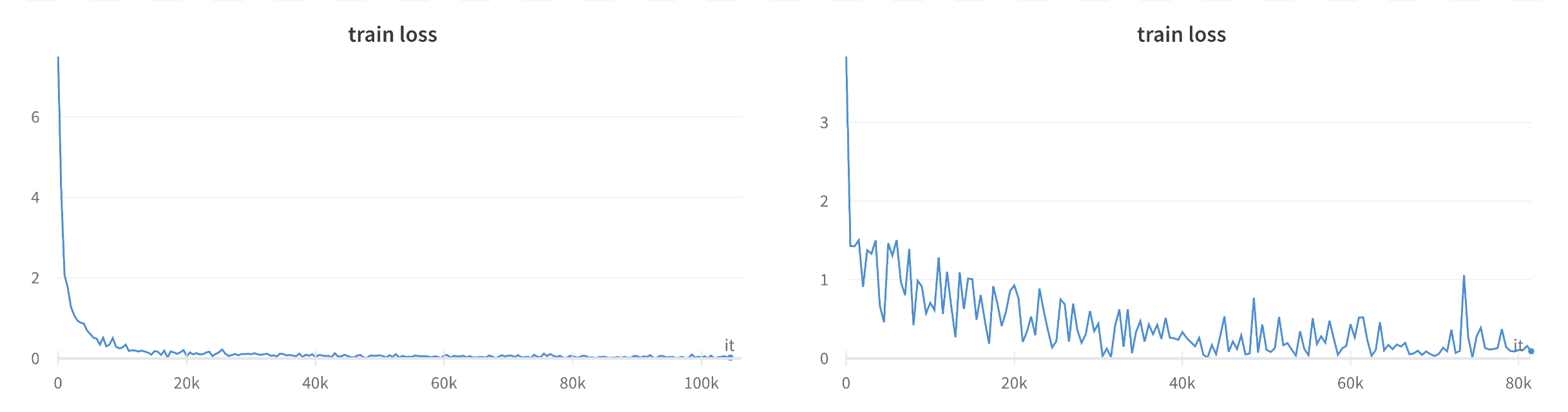}
\caption{
Plot showing convergence of Resnet12 on a high-diversity benchmark (hdb9-cavdo). The left plot depicts the training loss curve for the pre-training algorithm and the right plot depicts the training loss curve for MAML.
}
\label{res12_hdb9_learning_curve_pt_maml}
\end{figure*}

\textbf{Training to Convergence:}
We show how our models were trained by providing some sample learning curves for pre-training and MAML in the following Figures 
\ref{res12_micod_learning_curve_pt_maml},
\ref{mds_learning_curve_resnet50},
\ref{res12_fc100_learning_curve_pt_maml},
\ref{res12_aircraft_learning_curve_pt_maml},
\ref{res12_hdb8_learning_curve_pt_maml}, 
\ref{res12_hdb9_learning_curve_pt_maml}
.

\begin{figure*}[h!]
\centering
\includegraphics[width=1.0\linewidth]{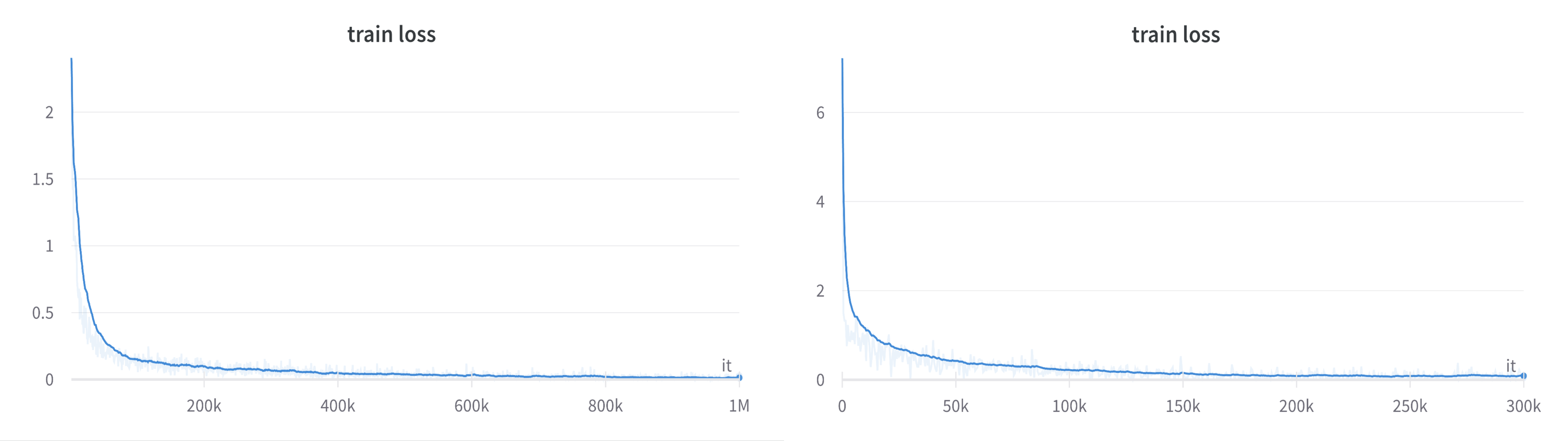}
\caption{
Plot showing convergence of Resnet12 on a high-diversity benchmark (MICOD). 
The left-most plot depicts the training loss curve for the pre-training algorithm, and the rightmost plot depicts the training loss curve for MAML.
}
\label{res12_micod_learning_curve_pt_maml}
\end{figure*}

\subsection{Decision Results using Confidence Intervals}
We present decision results obtained if we used traditional confidence intervals for analysis in Tables \ref{tab:comparison3}, \ref{tab:comparison4}, \ref{1pct_es_fo_lowdiv}, \ref{tab:comparison_1}, \ref{tab:comparison_2}, \ref{1pct_es_ho_lowdiv}, \ref{tab:comparison1}, \ref{tab:comparison2}, \ref{1pct_es_highdiv}.

\begin{table}[h]
\centering
\caption{
\textbf{Performance comparison between pre-training and (FO) MAML using confidence intervals for low-diversity benchmarks.} 
These performance comparison experiments were conducted using a batch size of 300.
}
\begin{tabular}{|c|c|c|}
\hline
Dataset & pt vs maml5 CI decision & pt vs maml10 CI decision \\ \hline
aircraft & H1 no diff & H1 no diff \\
flower & H1 pt & H1 pt \\
dtd & H1 pt & H1 pt \\
delaunay & H1 pt & H1 pt \\
cubirds & H1 maml5 & H1 maml10 \\
cifar-fs & H1 maml5 & H1 maml10 \\
fc100 & H1 maml5 & H1 maml10 \\
mini-imagenet & H1 pt & H0 no diff \\
omniglot & H0 no diff & H0 no diff \\
tiered-imagenet & H0 no diff & H0 no diff \\
vggair & H0 no diff & H0 no diff \\
vggdtd & H0 no diff & H0 no diff \\
\hline
\end{tabular}
\label{tab:comparison3}
\end{table}

\begin{table}[h]
\centering
\caption{
\textbf{Performance comparison between pre-training and (FO) MAML using confidence intervals for low-diversity benchmarks with a 1\% overlap threshold.}
These performance comparison experiments were conducted using a batch size of 300. 
}
\begin{tabular}{|c|c|c|}
\hline
Dataset & pt vs maml5 decision (1\% overlap) & pt vs maml10 decision (1\% overlap) \\ \hline
aircraft & H0 no diff & H0 no diff \\
flower & H1 pt & H1 pt \\
dtd & H1 pt & H1 pt \\
delaunay & H1 pt & H1 pt \\
cubirds & H1 maml5 & H1 maml10 \\
cifar-fs & H0 no diff & H1 maml10 \\
fc100 & H0 no diff & H0 no diff \\
mini-imagenet & H1 pt & H0 no diff \\
omniglot & H0 no diff & H0 no diff \\
tiered-imagenet & H0 no diff & H0 no diff \\
vggair & H0 no diff & H0 no diff \\
vggdtd & H0 no diff & H0 no diff \\
\hline
\end{tabular}
\label{tab:comparison4}
\end{table}

\begin{table}[h]
\centering
\caption{1\% effect sizes for performance comparison between pre-training and (FO) MAML for low-diversity benchmarks.}
\begin{tabular}{|c|c|c|}
\hline
Dataset & pt vs maml5 1\% ES & pt vs maml10 1\% ES \\ \hline
aircraft & 0.100 & 0.109 \\ 
flower & 0.171 & 0.195 \\
dtd & 0.122 & 0.125 \\ 
delaunay & 0.106 & 0.122 \\
cubirds & 0.135 & 0.133 \\ 
cifar-fs & 0.114 & 0.113 \\ 
fc100 & 0.121 & 0.117 \\ 
mini-imagenet & 0.123 & 0.117 \\ 
omniglot & 0.789 & 0.727 \\ 
tiered-imagenet & 0.112 & 0.113 \\ 
vggair & 0.059 & 0.059 \\ 
vggdtd & 0.056 & 0.056 \\
\hline
\end{tabular}
\label{1pct_es_fo_lowdiv}
\end{table}

\begin{table}[h]
\centering
\caption{
\textbf{Performance comparison between pre-training and (HO) MAML using confidence intervals for low-diversity benchmarks.}
These performance comparison experiments were conducted using a batch size of 300. 
}
\begin{tabular}{|c|c|c|}
\hline
Dataset & pt vs maml5 CI decision & pt vs maml10 CI decision \\ \hline
aircraft & H1 maml5 & H1 maml10 \\
flower & H1 pt & H1 pt \\
dtd & H1 pt & H1 pt \\
delaunay & H1 pt & H1 pt \\
cubirds & H1 maml5 & H1 maml10 \\
cifar-fs & H1 maml5 & H1 maml10 \\
fc100 & H1 maml5 & H1 maml10 \\
mini-imagenet & H0 no diff & H0 no diff \\
omniglot & H1 pt & H1 pt \\
tiered-imagenet & H1 pt & H1 pt \\
\hline
\end{tabular}
\label{tab:comparison_1}
\end{table}

\begin{table}[h]
\centering
\caption{
\textbf{Performance comparison between pre-training and (HO) MAML using confidence intervals for low-diversity benchmarks with a 1\% overlap threshold.}
These performance comparison experiments were conducted using a batch size of 300. 
}
\begin{tabular}{|c|c|c|}
\hline
Dataset & pt vs maml5 CI decision (1\% overlap) & pt vs maml10 CI decision (1\% overlap) \\ \hline
aircraft & H1 maml5 & H1 maml10 \\
flower & H0 no diff & H1 pt \\
dtd & H1 pt & H1 pt \\
delaunay & H1 pt & H1 pt \\
cubirds & H1 maml5 & H1 maml10 \\
cifar-fs & H1 maml5 & H1 maml10 \\
fc100 & H1 maml5 & H1 maml10 \\
mini-imagenet & H0 no diff & H0 no diff \\
omniglot & H0 no diff & H0 no diff \\
tiered-imagenet & H0 no diff & H1 pt \\
\hline
\end{tabular}
\label{tab:comparison_2}
\end{table}

\begin{table}[h]
\centering
\caption{1\% effect sizes for performance comparison between pre-training and (HO) MAML for low-diversity benchmarks.}
\begin{tabular}{|c|c|c|}
\hline
Dataset & pt vs maml5 1\% ES & pt vs maml10 1\% ES \\ \hline
aircraft & 0.094 & 0.100 \\ 
flower & 0.201 & 0.204 \\ 
dtd & 0.125 & 0.126 \\
delaunay & 0.116 & 0.115 \\ 
cubirds & 0.145 & 0.145 \\
fc100 & 0.113 & 0.113 \\ 
cifar-fs & 0.118 & 0.113 \\ 
mini-imagenet & 0.120 & 0.123 \\ 
omniglot & 0.655 & 0.762 \\ 
tiered-imagenet & 0.116 & 0.080 \\ 
\hline
\end{tabular}
\label{1pct_es_ho_lowdiv}
\end{table}

\begin{table}[h]
\centering
\caption{
\textbf{Performance comparison between PT and MAML using confidence intervals for high-diversity benchmarks.}
These performance comparison experiments were conducted using a batch size of 300. 
}
\begin{tabular}{|c|c|c|}
\hline
Dataset & pt vs maml5 CI decision & pt vs maml10 CI decision \\ \hline
hdb6-afdo & H0 no diff & H0 no diff \\
hdb7-afto & H0 no diff & H0 no diff \\
hdb8-cado & H0 no diff & H0 no diff \\
hdb9-cavdo & H0 no diff & H0 no diff \\
hdb10-micova & H1 maml5 & H1 maml10 \\
\hline
\end{tabular}
\label{tab:comparison1}
\end{table}

\begin{table}[h]
\centering
\caption{
\textbf{Results of performance comparison between PT and MAML using confidence intervals for high-diversity benchmarks with a 1\% overlap threshold.}
These performance comparison experiments were conducted using a batch size of 300. 
}
\begin{tabular}{|c|c|c|}
\hline
Dataset & pt vs maml5 CI decision (1\% overlap) & pt vs maml10 CI decision (1\% overlap) \\ \hline
hdb6-afdo & H0 no diff & H0 no diff \\
hdb7-afto & H0 no diff & H0 no diff \\
hdb8-cado & H0 no diff & H0 no diff \\
hdb9-cavdo & H0 no diff & H0 no diff \\
hdb10-micova & H1 maml5 & H1 maml10 \\
\hline
\end{tabular}
\label{tab:comparison2}
\end{table}

\begin{table}[h]
\centering
\caption{1\% effect sizes for performance comparison between pre-training and (FO) MAML for high-diversity benchmarks.}
\begin{tabular}{|c|c|c|}
\hline
Dataset & pt vs maml5 1\% ES & pt vs maml10 1\% ES \\ \hline
hdb6-afdo & 0.057 & 0.059 \\
hdb7-afto & 0.050 & 0.051 \\
hdb8-cado & 0.051 & 0.053 \\
hdb9-cavdo & 0.054 & 0.054  \\
hdb10-micova & 0.056 & 0.057  \\
\hline
\end{tabular}
\label{1pct_es_highdiv}
\end{table}

When comparing PT (pre-training) and MAML using confidence intervals, our experiments indicate that MAML and PT tend to perform equivalently under high-diversity benchmarks, while MAML and PT perform differently (either MAML outperforming or underperforming PT) under lower-diversity benchmarks.

\begin{figure*}[h!]
\centering
\includegraphics[width=1.0\linewidth]{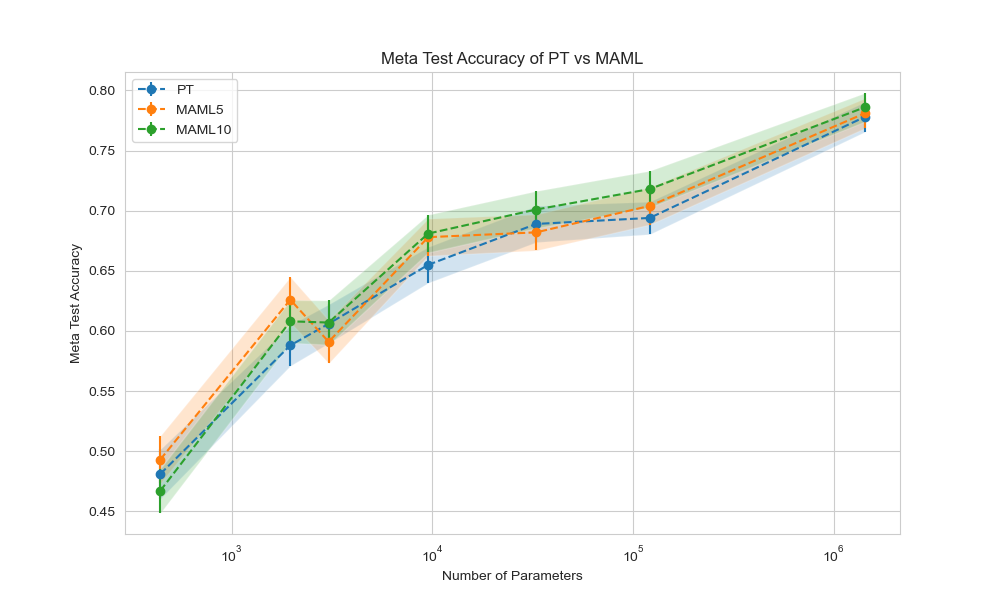}
\caption{ 
\textbf{Shows
how meta-test accuracy between PT and MAML(5,10) intersects in the high diversity dataset MICOD. 
}
However, on average MAML(5,10) performs better than PT.
This supports our main hypothesis because: 1. MAML is better than PT in the high diversity regime but 2. The difference is marginal, as shown by the confidence intervals being close. 
}
\label{fig:micod_cis}
\end{figure*}

\begin{figure*}[h!]
\centering
\includegraphics[width=1.0\linewidth]{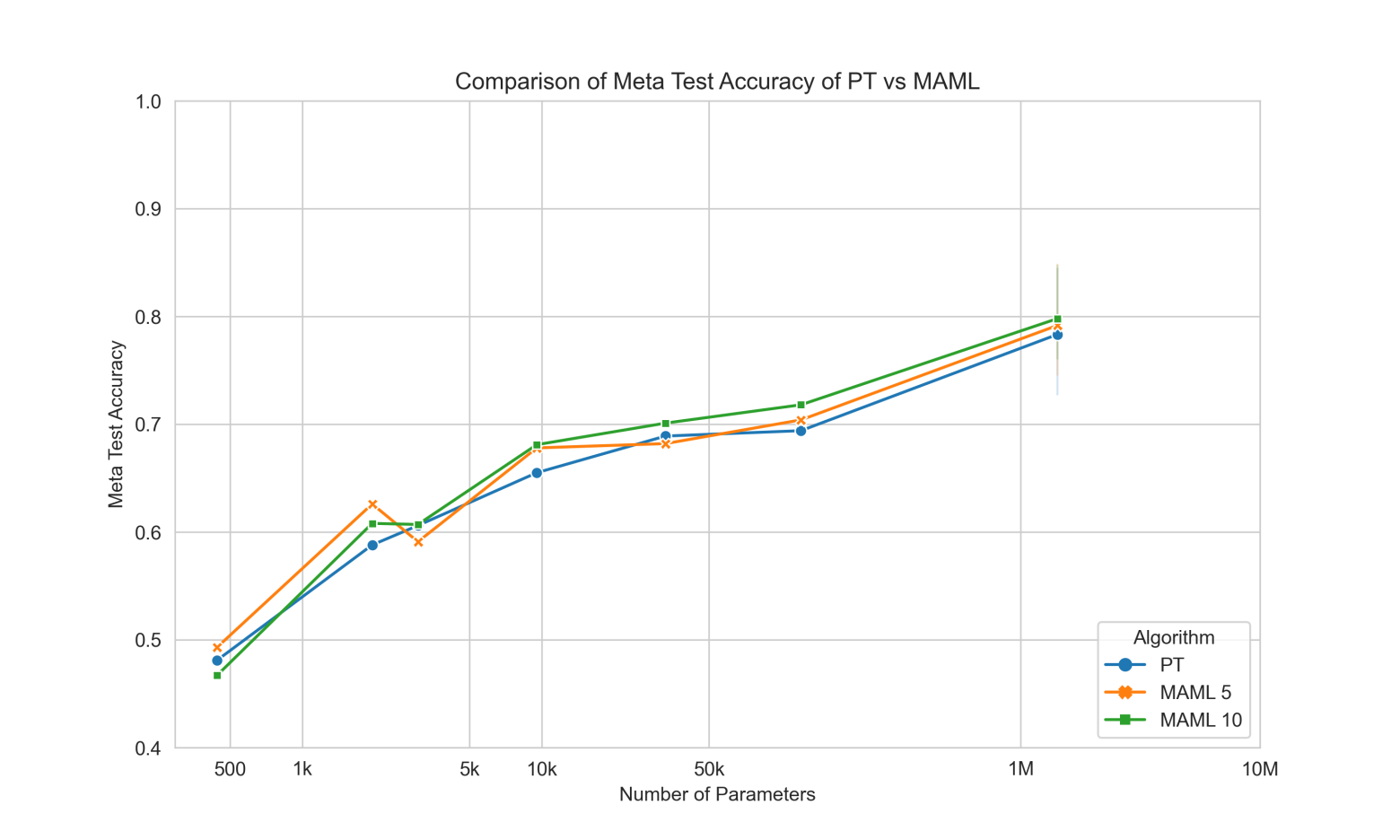}
\caption{ 
\textbf{ MAML(5,10) outperforms PT in the high diversity dataset MICOD. 
}
Same as Figure \ref{fig:micod_cis} but without confidence intervals. This supports our main hypothesis as it demonstrates that MAML is marginally better than PT in the high diversity regime.
}
\label{fig:micod_without_cis}
\end{figure*}

Figures \ref{fig:micod_cis} and \ref{fig:micod_without_cis} shows how average MAML(5,10) performs better than PT.
This supports our main hypothesis because 1. MAML is better than PT in the high diversity regime but 2. The difference is marginal, as shown by the confidence intervals being close. 

\subsection{L2 model norms and validation loss curves suggest that MAML has less meta-overfitting than PT}
\begin{figure*}[h!]
\centering
\includegraphics[width=1.0\linewidth]{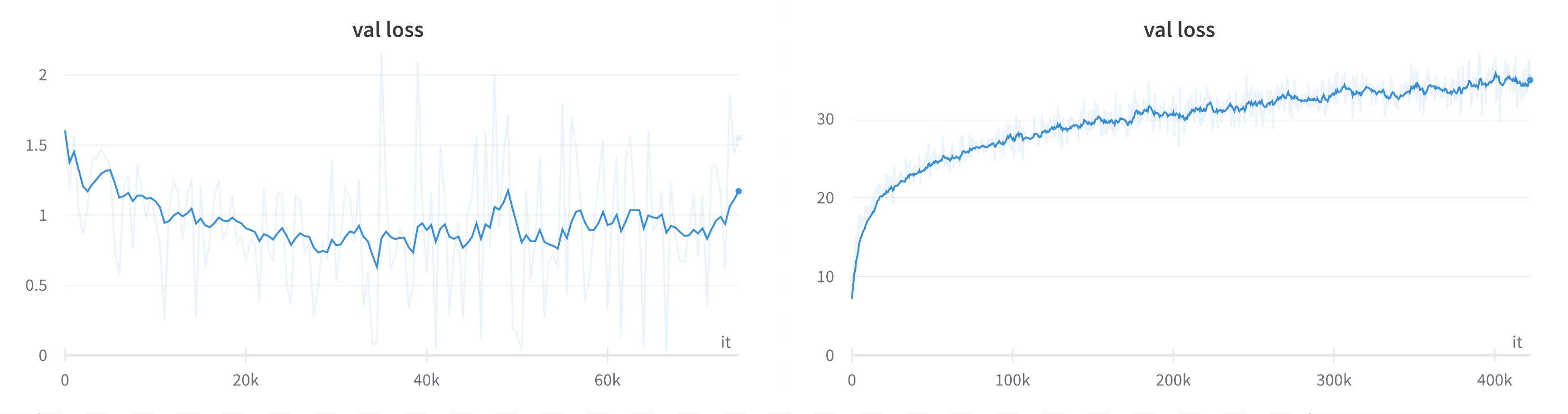}
\caption{
\textbf{On a high-diversity dataset, the validation loss of MAML stays relatively unchanged over time, while the validation loss of PT increases over time, suggesting that MAML has less meta-overfitting than PT.} The left plot depicts the validation loss curve for the MAML algorithm and the right plot depicts the validation loss curve for the PT algorithm, both on the high-diversity hdb8-cado dataset.
}
\label{hdb8_maml_vs_sl_val_loss}
\end{figure*}

\begin{figure*}[h!]
\centering
\includegraphics[width=1.0\linewidth]{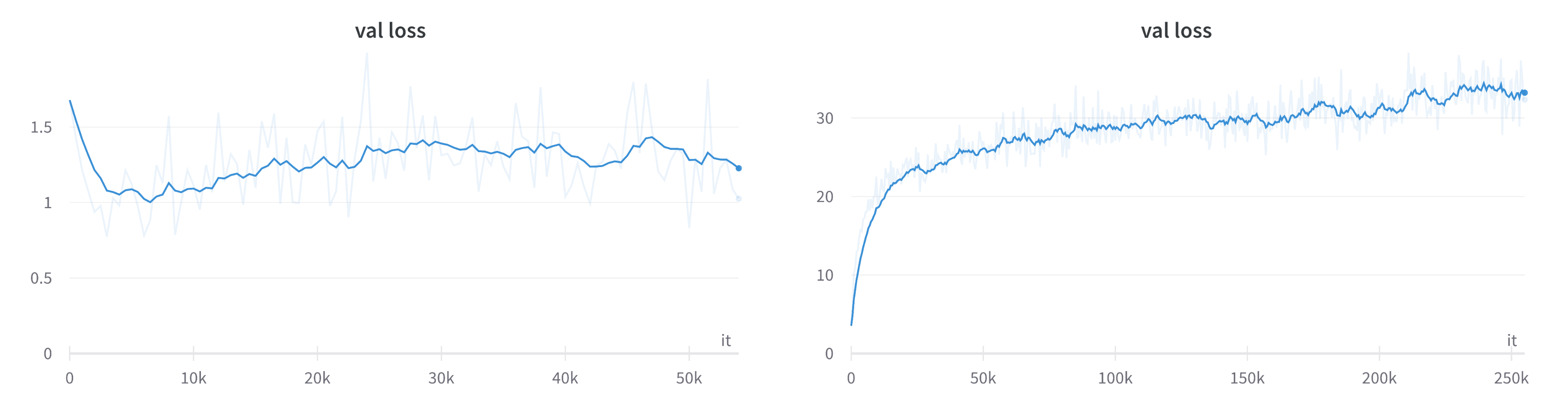}
\caption{
\textbf{On a low-diversity dataset, the validation loss of MAML stays relatively unchanged over time, while the validation loss of PT increases over time, suggesting that MAML has less meta-overfitting than PT.} The left plot depicts the validation loss curve for the MAML algorithm and the right plot depicts the validation loss curve for the PT algorithm, both on the low-diversity DTD dataset.
}
\label{dtd_maml_vs_sl_val_loss}
\end{figure*}

We demonstrate evidence that may suggest that MAML has less overfitting than PT, both via MAML and PT validation loss curves (see Figures \ref{hdb8_maml_vs_sl_val_loss} and \ref{dtd_maml_vs_sl_val_loss}), as well as the L2 model norms of trained MAML and PT models (see Tables \ref{l2_norm_maml_vs_pt_fo}, \ref{l2_norm_maml_vs_pt_ho}, \ref{l2_norm_maml_vs_pt_hdb}).

\begin{table}[h]
\centering
\caption{\textbf{The L2 norm of a trained first-order MAML model is less than the L2 norm of a trained PT model for each low-diversity benchmark, suggesting that MAML has less meta-overfitting than PT.}}
\begin{tabular}{|c|c|c|}
\hline
Dataset & L2 model norm (MAML) & L2 model norm (PT)  \\ \hline
cifar-fs & 851.012 & 9813.269 \\
fc100 & 919.964 & 8302.170 \\ 
omniglot & 663.715 & 4676.182 \\
mini-imagenet & 930.304 & 5776.625 \\
tiered-imagenet & 926.896 & 11097.097 \\
\hline
\end{tabular}
\label{l2_norm_maml_vs_pt_fo}
\end{table}

\begin{table}[h]
\centering
\caption{\textbf{The L2 norm of a trained higher-order MAML model is less than the L2 norm of a trained PT model for each low-diversity benchmark, suggesting that MAML has less meta-overfitting than PT.}}
\begin{tabular}{|c|c|c|}
\hline
Dataset & L2 model norm (MAML) & L2 model norm (PT)  \\ \hline
dtd & 2949.312 & 5548.194 \\
tiered-imagenet & 731.893 & 11097.097 \\
omniglot & 594.381 & 4676.182 \\
fc100 & 758.372 & 8302.170 \\
delaunay & 2810.343 & 4856.295 \\
aircraft & 1517.484 & 6144.078 \\
cifar-fs & 725.017 & 9813.269 \\
mini-imagenet & 752.201 & 5776.625 \\
cubirds & 3127.715 & 5252.014 \\
flower & 2333.556 & 7307.350 \\
\hline
\end{tabular}
\label{l2_norm_maml_vs_pt_ho}
\end{table}

\begin{table}[h]
\centering
\caption{\textbf{The L2 norm of a trained higher-order MAML model is less than the L2 norm of a trained PT model for each high-diversity benchmark, suggesting that MAML has less meta-overfitting than PT.}}
\begin{tabular}{|c|c|c|}
\hline
Dataset & L2 model norm (MAML) & L2 model norm (PT)  \\ \hline
hdb6-afdo & 2426.827 & 7699.514 \\
hdb7-afto & 2598.023 & 3573.5355 \\
hdb8-afdo & 3441.040 & 8072.174 \\
hdb9-cavdo & 3997.130 & 7919.635 \\
hdb10-micova & 3810.151 & 7757.541 \\
\hline
\end{tabular}
\label{l2_norm_maml_vs_pt_hdb}
\end{table}

\end{document}